\documentclass[journal]{IEEEtran}

\usepackage[utf8]{inputenc} 
\usepackage[T1]{fontenc}    
\usepackage{hyperref}       
\usepackage{url}            
\usepackage{booktabs}       
\usepackage{amsfonts}       
\usepackage{nicefrac}       
\usepackage{microtype}      

\usepackage{graphicx}
\usepackage{amsmath,amssymb} 
\usepackage{color,xcolor}
\usepackage[ruled,vlined]{algorithm2e}
\usepackage{multirow}

\usepackage{times}
\usepackage{epsfig}

\usepackage{subfigure}
\usepackage{lineno}
\usepackage{algpseudocode}  

\newcommand{\eg}{\textit{e.g.}}
\newcommand{\ie}{\textit{i.e.}}
\newcommand{\etal}{\textit{et al.}}

\newcommand{\Sdomain}{\mathcal{S}}
\newcommand{\Tdomain}{\mathcal{T}}

\title{Reciprocal Normalization for Domain Adaptation}

\author{Zhiyong~Huang, Kekai~Sheng, Ke Li, Jian Liang, Taiping Yao,  \\  Weiming~Dong,~\IEEEmembership{Member,~IEEE,}, Dengwen Zhou, Xing Sun,
\IEEEcompsocitemizethanks{
\IEEEcompsocthanksitem Z. Huang and D. Zhou are with School of Control and Computer Engineering, North China Electric Power University. E-mail: \{1182227193, zdw\}@ncepu.edu.cn.
\IEEEcompsocthanksitem K. Sheng, K. Li, T. Yao, and X. Sun are with Youtu lab, Tencent, Shanghai, China. E-mail: \{saulsheng, tristanli, taipingyao, winfredsun\}@tencent.com.
\IEEEcompsocthanksitem W. Dong and J. Liang are with NLPR, Institute of Automation, Chinese Academy of Sciences and School of Artificial Intelligence, University of Chinese Academy of Sciences, Beijing, China. E-mail: weiming.dong, @ia.ac.cn, liangjian92@gmail.com.
}
}

\begin{document}

\markboth{IEEE TRANSACTIONS ON IMAGE PROCESSING, VOL. XXX, 2021}%
{Huang \MakeLowercase{\textit{et al.}}: Reciprocal Normalization for Domain Adaptation}

\maketitle

\newcommand{\dps}{\mathbf{X}_{ncij}}
\newcommand{\SOTA}{\textit{state-of-the-art}}

\newcommand{\myparagraph}{\vspace{-4mm}\paragraph}

\newcommand{\tabincell}[2]{\begin{tabular}{@{}#1@{}}#2\end{tabular}}

\renewcommand{\algorithmicrequire}{\textbf{Input:}}
\renewcommand{\algorithmicensure}{\textbf{Ouput:}}

\begin{abstract}
Batch normalization (BN) is widely used in modern deep neural networks, which has been shown to represent the domain-related knowledge, and thus is ineffective 
for cross-domain tasks like unsupervised domain adaptation (UDA).
Existing BN variant methods aggregate source and target domain knowledge in the same channel in normalization module. However, the misalignment between the features of corresponding channels across domains often leads to a sub-optimal transferability.
In this paper, we exploit the cross-domain relation and propose a novel normalization method, Reciprocal Normalization (RN).
Specifically, RN first presents a Reciprocal Compensation (RC) module to acquire the compensatory for each channel in both domains based on the cross-domain channel-wise correlation.
%
Then RN develops a Reciprocal Aggregation (RA) module to adaptively aggregate the feature with its cross-domain compensatory components.
As an alternative to BN, RN is more suitable for UDA problems and can be easily integrated into popular domain adaptation methods.
Experiments show that the proposed RN outperforms existing normalization counterparts by a large margin and helps \SOTA\ adaptation approaches achieve better results.
The source code is available on \url{https://github.com/Openning07/reciprocal-normalization-for-DA}.
\end{abstract}


\section{Introduction}
\label{sec:introduction}
Unsupervised domain adaptation (UDA)~\cite{ganin2015DANN,long2018CDAN,kang2019CAN,liang2020shot,liang2021shot,zuo2021attention} aims to transfer the knowledge learned from the labeled source domain to the unlabeled target domain.
It has been widely applied in classification~\cite{long2017JAN}, detection~\cite{xu2020cross}, and segmentation~\cite{zhou2020affinity}.
Technically, besides prevailing feature alignment~\cite{ganin2015DANN, long2018CDAN, Chen_2019_CVPR} and pixel-level image translation~\cite{Murez_2018_CVPR,Pizzati_2020_WACV}, to enhance the feature transferability and learn domain-specific knowledge better, many researchers (\eg, ~\cite{li2017AdaBN,cariucci2017autodial,wang2019TN,chang2019domain}) focus on improving the feature normalization module in deep neural networks (DNNs) to narrow the domain gap.

\begin{figure}
    \centering

    \centering
    \includegraphics[width=\linewidth]{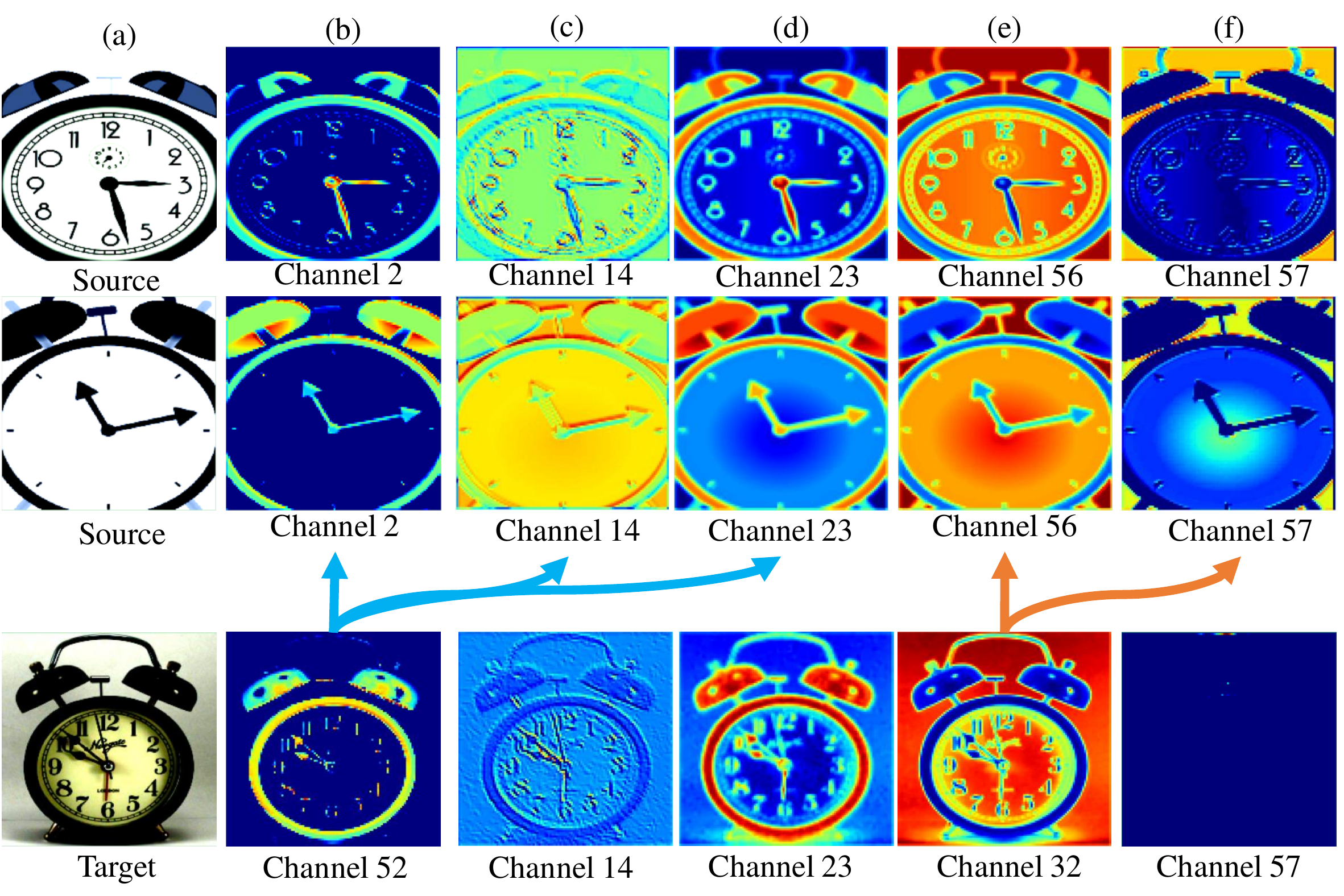}
    \caption{The visualization of feature maps from the first ReLU layer of ResNet-50~\cite{he2016resnet} on the UDA task Clipart (1st and 2nd rows) $\to$ Art (3rd row), which is trained with CDAN~\cite{long2018CDAN}.
    (a) $3$ images of Alarm Clock. 
    (d) is the similar pattern at the same channel.
    (c) and (f) are different patterns at the same channels. 
    (b) and (e) are the similar patterns at the different channels. 
   }
    \label{fig:channel_diff}
    \vspace{-2mm}
\end{figure}

Technically, batch normalization (BN)~\cite{ioffe2015BN} is a powerful approach to alleviate the internal covariate shift and has been widely used in DNNs, \eg, ResNet-50~\cite{he2016resnet}. 
Nevertheless, recent research works~\cite{chang2019domain,wang2019TN} point out that BN suffers from losing domain-specific information in the UDA scenario, because sharing the mean and variance for the two domains are inappropriate~\cite{wang2019TN}.
To compensate for the deficiency of BN, several methods are proposed to preserve the domain-specific knowledge~\cite{li2017AdaBN,cariucci2017autodial,wang2019TN,chang2019domain}.
AdaBN~\cite{li2017AdaBN} uses different domain statistics for the two domains.
However, only employing the target statistics in the inference can lose the information of the source domain.  
To merge the information of different domains, AutoDIAL~\cite{cariucci2017autodial} fuses domain statistics channel by channel using a shared weight parameter for each channel.
TN~\cite{wang2019TN} proposes a channel attention mechanism to highlight the channels with high transferability to further focus on the important information.

The aforementioned methods reinforce UDA by aggregating the domain knowledge extracted from the corresponding channels.
For different examples from the same domain, the learned patterns are likely to be captured by the same channel (see the upper and middle rows in Fig.~\ref{fig:channel_diff}).
When encountering cross-domain scenarios, we observe that the same or similar patterns cannot always be captured by the same channel, however, which is always ignored by existing UDA methods. 
As illustrated in Figs.~\ref{fig:channel_diff} (c) and (f), different patterns are captured by the same channels across domains.
%
Thus, merging the domain knowledge of corresponding channels across domains in ~\cite{cariucci2017autodial} can inevitably lose domain-specific information and lead to sub-optimal UDA performance.
Another important observation is that similar patterns from different domains are likely to exist in the non-corresponding channels (\eg, Figs.~\ref{fig:channel_diff}(b) and \ref{fig:channel_diff}(e)). 
Moreover, the shareable patterns at non-corresponding channels across domains are not just the one-to-one relationship, as shown in Fig.~\ref{fig:channel_diff} (the orange and blue arrows). 
%
Therefore, adaptively considering the correlations of all cross-domain channels is crucial to break through the bottleneck in DA architectures.

\begin{figure*}
    \centering
    \includegraphics[width=0.96\linewidth]{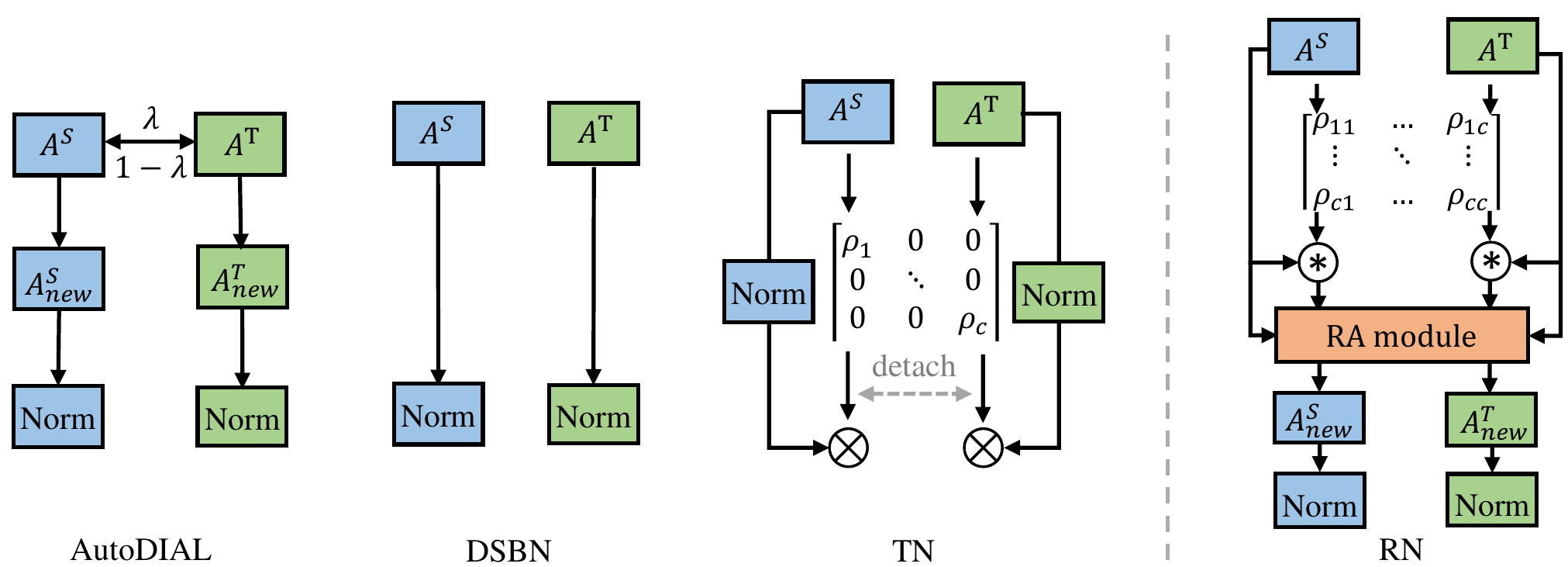}
    \caption{The differences between our RN and other typical UDA normalization technologies. 
    The $A^S$ and $A^T$ denote the data statistics of source and target domains, respectively. Specifically, DSBN~\cite{chang2019domain} adopts totally separated normalization methods for both source and target domains. 
    AutoDIAL~\cite{cariucci2017autodial} and TN~\cite{wang2019TN} consider the correlations of corresponding cross-domain channels to enhance the transferability. While our RN is able to capture long-range correlations between cross-domain channels. The probability of TN is detached from calculation graph. Our RN utilizes a Reciprocal Aggregation (RA) module to adaptively aggregate both source and target information.}
    \label{fig:norm_com}
\end{figure*}

Building on the observations and deductions above, in this paper, we propose a novel Reciprocal Normalization (RN) scheme for unsupervised domain adaptation. 
Fig.~\ref{fig:norm_com} illustrates the key differences between existing UDA normalization techniques and our RN.
In contrast to the local behavior of BN and its variants towards domain adaptation, the proposed RN is able to capture long-range correlations directly by computing interactions between any two cross-domain channels and then conducts reciprocity between domains during normalization.
Specifically, we firstly present a reciprocal compensation (RC) module to acquire the compensatory of each source/target channel for the counterpart in the target/source domain by modeling the correlation of any two cross-domain channels. 
For efficient reciprocity and effective domain alignment,
we then develop a Reciprocal Aggregation (RA) module to adaptively aggregate the feature with its cross-domain compensatory component. Put RC and RA together, we propose RN to boost the performance of various domain adaptation tasks.
 
In summary, our main contributions are three-fold:
\begin{itemize}
    \setlength{\itemsep}{0pt}
    \setlength{\parsep}{0pt}
    \item We propose a novel RN scheme for domain adaptation to address the issue of channel misalignment across domains and get better performance on the target domain. 
    
    \item The proposed RN structurally aligns the source and target domains by conducting reciprocity across domains.
    Besides being a plug-and-play module, RN can be also integrated with other domain adaptation methods to achieve better results.
    
    \item Experiments on three benchmarks (ImageCLEF-DA, Office-Home, and VisDA-C) and various DA scenarios (closed-set DA, partial-set DA, and multi-source DA). Extensive results indicate that our RN outperforms existing normalization methods and effectively improves the performance of \SOTA\  domain adaptation approaches in various scenarios.
    
\end{itemize}

\section{Related Work}

\subsection{Domain Adaptation}
Existing approaches mainly focus on loss function design or network design.
Technically, the loss function design usually starts from two directions.
\romannumeral1) To match all statistics of the two domains to minimize cross-domain distribution discrepancy:
DDC~\cite{tzeng2014deep} and DAN~\cite{long2015DAN} employ Maximum
Mean Discrepancy (MMD)~\cite{greron2007mmd} to measure and reduce the discrepancy of source and target domains; 
JAN~\cite{long2017JAN} utilizes Joint Maximum Mean Discrepancy to combine adversarial learning with MMD;
SWD \cite{lee2019sliced} introduces Sliced Wasserstein Distance and CAN~\cite{kang2019CAN} leverages Contrastive Domain Discrepancy to find a better measure of the domain discrepancy.
\romannumeral2) To introduce domain discriminators and exploit adversarial learning to encourage domain confusion:
DANN~\cite{ganin2015DANN} introduces domain adversarial loss to learn domain-invariant representations;
ADDA~\cite{tzeng2017ADDA} combines adversarial learning with discriminative feature learning via adopting asymmetric feature extractors for each domain;
CDAN~\cite{long2018CDAN} employs a conditional domain-adversarial paradigm to train an adversarial adaptation model.
More recently, advanced loss functions (\eg, BSP~\cite{chen2019BSP}, IAA~\cite{jiang2020implicit}, BNM~\cite{cui2020BNM}, and SRDC~\cite{tang2020SRDC}), learning schemes~\cite{mancini2018DAlayers,sankaranarayanan2018GTA,shu2018dirt,wang2020SPL,zhang2020APL,xiao2021DWL} and new network designs (\eg, TN~\cite{wang2019TN}, DCAN~\cite{li2020DCAN}, and BCDM~\cite{Li21BCDM}) are proposed for better performance on target domain.

However, all these existing methods overlook the the misalignment between the features of corresponding channels across domains, which often leads to a sub-optimal DA performance.
Additionally, as a general method, our work is able to benefit many unsupervised domain adaptation scenarios including vanilla closed-set UDA, partial-set DA (PDA), and multi-source DA (MSDA).


\subsection{Normalization Techniques}
It is widely applied in CNNs to make them learn faster, more stable, and increase their generalization ability~\cite{tseng2019FWT,wang2019TN,du2020metanorm}.
Representative methods include BN~\cite{ioffe2015BN}, Layer Normalization (LN)~\cite{ba2016LN}, Adaptive BN (AdaBN)~\cite{li2017AdaBN}, Group Normalization (GN)~\cite{wu2018GN}, Switchable Normalization (SN)~\cite{luo2018SN}, TaskNorm~\cite{bronskill2020tasknorm}, EvoNorm~\cite{liu2020EvoNorm}, Meta-Norm~\cite{du2020metanorm}, and Representation Normalization~\cite{gao2021representative}.
For better domain adaptation, researchers have devised novel designs to mitigate the shortcomings in BN.
AdaBN~\cite{li2017AdaBN} uses the statistics of source domain during training and those of target domain during evaluation, respectively.
AutoDIAL~\cite{cariucci2017autodial} integrates the statistics of two domains channel by channel in order to align the source and target feature distributions.
Domain Specific BN (DSBN)~\cite{chang2019domain} normalizes the source and target representations completely individually, including affine parameters. 
Transferable Normalization (TN)~\cite{wang2019TN} utilizes the statistics of two domains to calculate corresponding channel attention, which are all detached from the computation graph.
ConvNorm~\cite{li2019ConvNorm} proposes an adaptation layer $\mathcal{A}$ to whiten and color source domain data, then $\mathcal{A}$ is fine-tuned on the target domain.
DWT~\cite{roy2019DWT} uses two co-variance matrices to whiten feature maps from source and target domains, respectively.
Particularly, DSBN~\cite{chang2019domain} and DWT~\cite{roy2019DWT} adopt totally separately normalize feature maps from source and target domains.
AutoDIAL~\cite{cariucci2017autodial} and TN~\cite{wang2019TN} consider the corresponding cross-domain channels to enhance the transferability.
They achieve promising progresses but neglect the misalignment between non-corresponding  channels across domains.

Different from these existing normalization counterparts, we focus on modeling the non-corresponding channels in CNNs for domain adaptation. 
In this paper, we propose a novel feature normalization method that facilitates domain alignment via conducting cross-domain reciprocity.

\section{Methodology}
In this section, we present the details of our RN.
We firstly revisit the BN and reformulate it for clear presentation (Section~\ref{method:preliminary}).
Then, we introduce RN to alleviate the misalignment between features across domains (Section~\ref{method:RN}).

\subsection{Revisiting Batch Normalization}
\label{method:preliminary}
Batch normalization (BN)~\cite{ioffe2015BN} is excellent in CNNs for many visual recognition tasks.
Technically, the BN layer firstly estimates the standardized features ($\ie$ with zero mean and unit standard deviation) at the channel dimension on the basis of mini-batch data, and then scales and shifts the standardized features by using a pair of learnable parameters $\gamma$ and $
\beta$.
Given the feature $x \in \mathbb{R}^{N \times C \times H \times W}$, the transformed $\widehat{x}$ is acquired through BN layer as:
\begin{equation}
    \widehat{x}^{(i)} = \gamma\hat{x} + \beta, \quad
    \hat{x} = \frac{x  - \mu}{\sqrt{(\sigma^{2})  + \epsilon}},
\end{equation}
where $\epsilon$ is a small constant for numerical stability. $\mu$ and $\sigma^{2}$ are the \emph{mean} and \emph{variance} statistics for each channel over a mini-batch respectively, and are defined as:
\begin{equation}
    \mu = \frac{1}{N}\sum^{N}_{n=1}x_{n},
    \quad
    \sigma^{2} = \frac{1}{N}\sum^{N}_{n=1}(x_{n} - \mu^{2}).
    \label{eq:mu_and_sigma}
\end{equation}
To obtain the accumulated statistics for the whole training data, the BN layer keeps running estimates towards the $\mu$ and $\sigma^{2}$ to obtain $\bar{\mu}$ and  $\bar{\sigma}^{2}$ during training phase:
\begin{equation}
    \begin{split}
    \bar{\mu}^{t+1} &= (1-\alpha) \bar{\mu}^{t} + \alpha \mu^{t}, \\
    (\bar{\sigma}^{t+1})^2 &= (1-\alpha)(\bar{\sigma}^{t})^{2} + \alpha (\sigma^t)^{2},
    \end{split}
\end{equation}
where $\alpha$ denotes the momentum and $t$ is the index of mini-batch data.
The estimated mean and variance will be used to normalize the features during inference phase.
In this way, the BN layer can successfully accelerate and stabilize training.
However, it is somewhat unreasonable to directly share the same mean and variance statistics between source and target domains since there exists a significant gap.

\begin{figure*}
    \centering
    \includegraphics[width=0.9\linewidth]{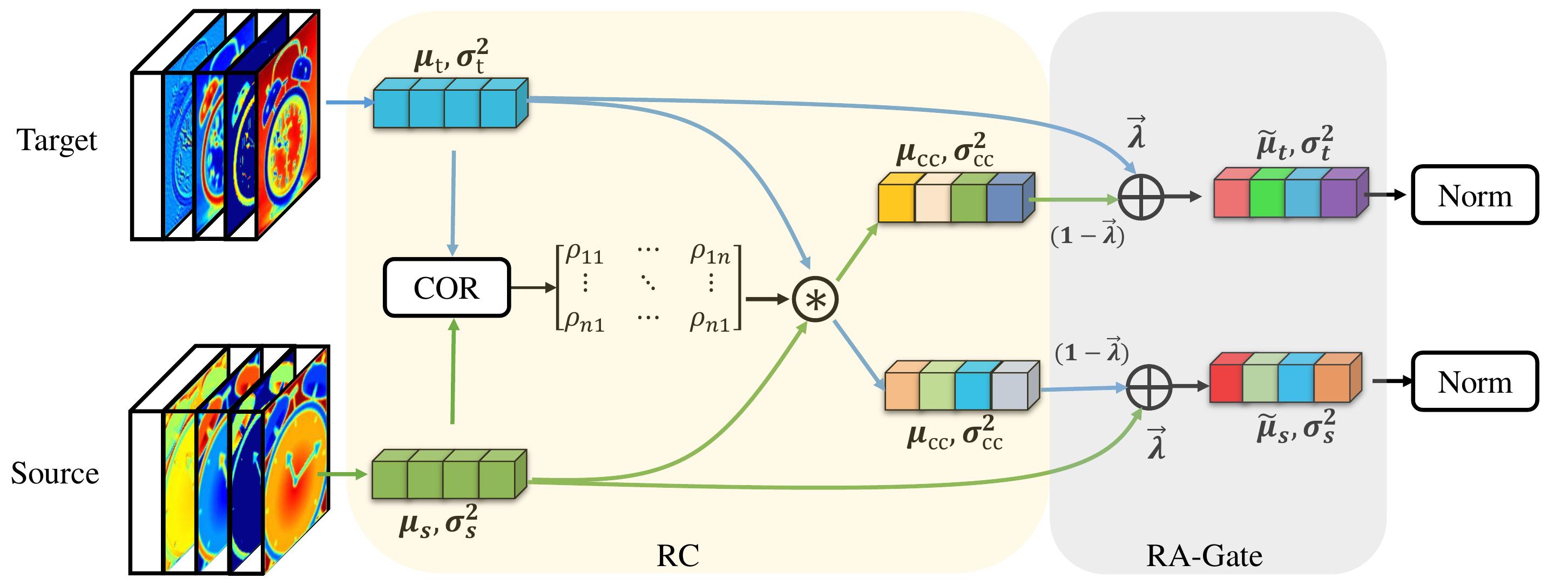}
    \caption{A main schematic diagram of RN. The blue \textcolor{blue}{$\to$}  and the red \textcolor{red}{$\to$} denote the target and the  source information flows, respectively. The ``\textit{COR}'' denotes calculating the correlations between cross-domain channels. $\Vec{\lambda}$ denotes the weight vector of reciprocal aggregation. }
    \label{fig:our_method}
\end{figure*}

\subsection{Reciprocity Normalization (RN)}
\label{method:RN}
Several methods have recently been proposed to address the limitation of BN, such as AdaBN~\cite{li2017AdaBN}, AutoDIAL~\cite{cariucci2017autodial}, DWT~\cite{roy2019DWT}, DSBN~\cite{chang2019domain}, and TN~\cite{wang2019TN}.
We illustrate the main differences between other typical UDA normalization techniques and our RN in Fig.~\ref{fig:norm_com}.
Generally, those methods all adopt separate normalization to avoid sharing exactly the same mean and variance.
%
However, such a mechanism suffers from another problem, \ie,  the misalignment of activations in the corresponding channel across domains, which sometimes leads to negative transfer. 
Due to the differences in background, style, distribution, \emph{e.t.c.}, between domains, it is intuitive that similar patterns of source and target domains are likely to be activated by non-corresponding cross-domain channels (\eg, Fig.~\ref{fig:channel_diff}(c) and Fig.~\ref{fig:channel_diff}(f)). 
As a result, simply normalizing source and target features separately may lose the domain information. 
Although AutoDIAL and TN care for the information of corresponding channels, they only partially alleviate the problem since they neglect the correlation between non-corresponding channels.

Motivated by the aforementioned observations and analyses, we propose a novel RN method for domain adaptation.
The main pipeline of RN is shown in Fig.~\ref{fig:our_method}. It consists of two main procedures: RC and RA. 
For a convenient and concise expression, we only present the reciprocity from the source domain to the target domain, and the other half of the corresponding operation is basically the same.  

\subsubsection{Reciprocal Compensation (RC)}
It models the relationship of any two channels across domains. 
The key insight is that similar patterns between domains are likely to be captured by not only the corresponding but also non-corresponding channels across domains (\ie, one-to-more relationship) when the domain shift is significant. 
We aim to fully consider any two cross-domain channels and then conduct reciprocity between domains.

Specifically, we first calculate the source ($s$) and target ($t$) statistics $\mu_s, \sigma^2_s$ and $\mu_t, \sigma^2_t$ via Equation~(\ref{eq:mu_and_sigma}).
To enable the channels with similar characteristics to have more correlation, we compute the correlation between any two channels via the negative $l_2$ distance:
\begin{equation}
    E_{i, j}^{\mu} = -(\mu_{i, t} - \mu_{j, s})^2, \quad
    E_{i, j}^{\sigma^2} = -(\sigma^2_{i, t} - \sigma^2_{j, s})^2,
\end{equation}
where $E_{i, j}$ denotes the correlation between the $i$-th channel of target domain and the $j$-th channel of source domain.
Below we use $E_{t \to s}^{\mu}$ and $E_{t \to s}^{\sigma^2}$ to denote the two correlation matrices.
In Section~\ref{Analysis}, we compare the results of some popular distance measures and find that $l_2$ distance performs the best, thus we choose $l_2$ distance as our default setting.

Then, to obtain the probabilistic weights of correlation between any two cross-domain channels, we normalize $E_{t \to s}^{\mu}$ and $E_{t \to s}^{\sigma^2}$ at the row dimension with softmax layer. The correlation score matrices $\rho_{t \to s}^{\mu}$ and $\rho_{t \to s}^{\sigma^2}$ can be computed respectively via:
\begin{equation}
    \begin{split}
    \rho_{t \to s}^{\mu} &= softmax(E_{t \to s}^{\mu}, \texttt{dim=1}), \\
    \rho_{t \to s}^{\sigma^2} &= softmax(E_{t \to s}^{\sigma^2}, \texttt{dim=1}),
    \end{split}
\end{equation}
where \texttt{dim=1} denotes the normalization of the matrices at the row dimension. In this way, we obtain the normalized correlation probability between each channel of target domain and all channels of source domain. 
This appears similar to TN~\cite{wang2019TN} that quantifies the transferability of corresponding channels across domains to calculate the channel attention weights.
However, TN neglects the correlation between non-corresponding channels across domains. 
Usually, the limitation of misalignment of channels can be partially mitigated by TN, but TN just puts a large emphasis on the corresponding channels with similar patterns and neglects to fully exploit the similar patterns in non-corresponding channels. It also leads to the loss of the domain information of corresponding channels with different patterns to a certain extent. 

Finally, the compensatory of each channel of target domains can be computed in the source domain space.
The compensatory  of $\mu_t$ and $\sigma^2_{t}$ can be obtained by:
\begin{equation}
    \begin{split}
    \mu_{t, cc} = \rho_{t \to s}^{\mu} \cdot \mu_{s}, \quad
    \sigma^2_{t, cc} = \rho_{t \to s}^{\sigma^2} \cdot \sigma^2_{s}.
    \end{split}
\end{equation}
Such calculation allows each compensatory  of channels to capture long-range correlations directly by conducting reciprocity among all cross-domain channels, including similar and complementary channels.

Furthermore, the global domain information is exploited by RC beyond the limit of the local receptive field of the convolutional kernel.

\subsubsection{Reciprocal Aggregation (RA)}
Although we have obtained the compensatory for each channel of the target domain in the source domain space, it is inappropriate to directly utilize $\mu_{cand}$ and $\sigma^2_{cand}$ to conduct the feature normalization since it may cause the loss of original domain-specific knowledge. The empirical results in Section~\ref{Analysis} also verify this judgment.
Thus, we aim to enable our module to adaptively learn the degree of reciprocity of domain information from various deep layers.
Particularly, AutoDIAL~\cite{cariucci2017autodial} directly integrates the statistics of two domains via one single 1-D parameter to endow the network with the ability to automatically align source and target domains.
We follow this strategy and develop RA to adaptively aggregate the matched compensatory  and the original domain statistics.
Specifically, we introduce the learnable gate parameters $g \in [0.5, 1]^C$:
\begin{equation}
    \begin{split}
    \widetilde{\mu}_{t} &= g^{\mu}_{t} * \mu_{t} + (1-g^{\mu}_{t}) * \mu_{t, cc}, \\
    \widetilde{\sigma}^2_{s} &= g^{\sigma^2}_{t} * \sigma^2_{t} + (1-g^{\sigma^2}_{t}) * \sigma^2_{t, cc},
    \end{split}
\end{equation}
where ``$*$'' denotes the Hadamard product. 
During training, $g$ is initialized as a unit vector so that RN performs the pure domain-specific normalization at the beginning of training, and then reduces the domain discrepancy via bridging the gaps between the source and target domains with $g$ updated progressively. 
Due to such an aggregation between each channel and its compensatory, the mutual domain information associated with individual channels can be emphasized accordingly.
Different from AutoDIAL~\cite{cariucci2017autodial} that directly mixes statistics of the two domains, RN uses RC to produce the information fed into aggregation
This scheme considers the correlation between any two cross-domain channels so as to contain more domain information.
Besides, AutoDIAL uses a single 1-D parameter to align all the domain statistics, which may be less effective and adaptive.
By contrast, the mean and variance are equipped with their own C-D parameters, endowing RN with the ability to adaptively learn where and how to conduct aggregation.
\begin{algorithm}[t]
    \small{
    \caption{The forward pass of RN during training}
    \begin{algorithmic} 
    \Require  Feature maps of source and target domains in a mini-batch: $\{x_{s}, x_{t}\} \in \mathbb{R}^{N \times C \times H \times W}$; learnable $g$ in RA: $\{ g_s^{\mu}$, $g_s^{\sigma^2}$, $g_t^{\mu}$, $g_t^{\sigma^2} \} \in [0.5, 1]^C$; learnable affine parameters: $\gamma$. $\beta$.
    \Ensure $\widehat{x}_s, \ \widehat{x}_t$
    \State Calculate  $\{ \mu_{s}, \mu_{t}, \sigma^2_s, \sigma^2_t \} \in \mathbb{R}^{C \times 1}$ \\
    For concise  expression, let $z \in \{ \mu, \sigma^2 \}$   
    \\ Calculate the correlation strength:
    \State  $ E_{t \to s}^{z}$ $\gets$ $ -(z_{i, t} - z_{j, s})^2$,
    \State  $ E_{s \to t}^{z} \gets (E_{t \to s}^{z})^T$
    \\ Obtain the probabilistic weights of correlation:
    \State 
     $\rho_{t \to s}^{z} \gets softmax(E_{t \to s}^{z}, \texttt{dim=1})$ \\
    $\rho_{s \to t}^{z} \gets softmax(E_{s \to t}^{z}, \texttt{dim=1})$
    \\ Obtain the candidates of each statistics:
    \State $z_{t, cc} \gets \rho_{t \to s}^{z} \cdot z_s$ \\
    $z_{s, cc} \gets \rho_{s \to t}^{z} \cdot z_t$ 
    \\ Reciprocal aggregation:
    \State $\Tilde{z_t} \gets g_t^z * z_t + (1-g_t^z) * z_{t, cc}$\\
    $\Tilde{z_s} \gets g_s^z * z_s + (1-g_s^z) * z_{s, cc}$
    \State  $\widehat{x_s} \gets \gamma \frac{x_{s}  - \tilde{\mu}_s}{\sqrt{\tilde{\sigma}^{2}_{s} + \epsilon}} + \beta$, \,
    $\widehat{x_t} \gets \gamma \frac{x_{t}  - \tilde{\mu}_t}{\sqrt{\tilde{\sigma}^{2}_{t} + \epsilon}} + \beta$
    \end{algorithmic}
    }
\end{algorithm}

\subsubsection{Separate Normalization}
Without loss of generality, we adopt the aggregated domain statistics to normalize the feature representations from source and target domains, separately. Akin to BN, we utilize affine parameters $\gamma$ and $\beta$ to re-scale and re-shift the normalized feature responses, where $\gamma$ and $\beta$ are shared in the two domains.
Here, we just present the normalization of target domain as follows:
\begin{equation}
    \widehat{x}^{(i)}_{t} = \gamma^{(i)}\hat{x}^{(i)}_{t} + \beta^{(i)},
    \quad
    \hat{x}^{(i)} = \frac{x^{(i)}_{t}  - \widetilde{\mu}^{(i)}}{\sqrt{\widetilde{\sigma}^{2(i)}_{t} + \epsilon}},
\end{equation}
where $\epsilon$ is a small constant to avoid divide-by-zero.
In this way, the domain-specific information can be well captured at the early training stage and the alignment between the source domain and the target domain can be progressively carried out via adaptive reciprocity. 

\subsubsection{Inference}
To reduce time cost at inference, we adopt a memory strategy similar to BN.
During training, RN keeps running estimates of its aggregated mean and variance of each domain, via exponential moving average with a hyper-parameter $\alpha$, which is given by:
\begin{equation}
    \begin{split}
    \bar{\mu}_{d}^{t+1} &= (1 - \alpha) \bar{\mu}_{d}^{t} + \alpha \widetilde{\mu}_{d}^t, \\
    (\bar{\sigma}^{t+1}_{d})^2 &= (1 - \alpha) \bar{\sigma}^{t}_{d} + \alpha  (\widetilde{\sigma}^t_{d})^2,
    \end{split}
\end{equation}
where $d \in \{s, t\}$, $\alpha$ is initialized to $0.1$, and the estimated aggregated statistics $\bar{\mu}_{t}$ and $\bar{\sigma}_{t}^2$ are used for the examples from target domain at inference. 
Such a strategy allows RN directly utilize the estimated domain statistics to normalize the examples during the evaluation without performing secondary calculations about RC and RA. 


\section{Experiments}
In this section, we evaluate the proposed RN on three benchmarks of three adaptation scenarios: vanilla closed-set UDA, partial-set DA (PDA), and multi-source DA (MSDA). We compare the performance of RN and the other existing normalization approaches, including their computation cost in training and inference stages. Besides, we conduct ablation study of the two modules (\ie, RC and RA) in our RN.
To better understand the rationale and the working mechanism of RN, we have some theoretical analyses based on quantitative results and feature visualization.
For additional implementation details, please refer to our Github project~\footnote{\url{https://github.com/Openning07/reciprocal-normalization-for-DA}.}.

\subsection{Setup}

\subsubsection{Datasets}
We experiment on three cross-domain benchmarks.
i) \textit{ImageCLEF-DA} is a small-scale dataset with $12$ classes shared by $3$ domains: Caltech-256 (C), ILSVRC 2012 (I), and Pascal VOC 2012 (P). We conduct experiments on all the $6$ transfer tasks.
ii) \textit{Office-Home}~\cite{venkateswara2017deep} is a medium-sized benchmark of $12$ adaptation tasks from $4$ domains: Artistic (Ar), Clip Art (Cl), Product (Pr), and Real-World (Rw). Each domain contains $65$ everyday object categories.
iii) \textit{VisDA-C}~\cite{peng2017visda} is a challenging large-scale benchmark of $12$-class synthesis-to-real adaptation task. The source domain contains $152K$ synthetic images generated by rendering 3D models, and the target domain has $55 K$ real object images.

\begin{table*}
    \centering
    \small{
    \setlength{\tabcolsep}{0.8mm}{
    \caption{Accuracy ($\%$) on Office-Home benchmark for ResNet-50-based UDA and PDA methods. The best results are in \textbf{bold}.}
    \label{tab:OfficeHome_results}
    \begin{tabular}{lccccccccccccc}
        \toprule
        \textbf{Closed-set UDA} & Ar$\to$Cl & Ar $\to$ Pr & Ar $\to$ Rw & Cl$\to$Ar & Cl$\to$Pr & Cl$\to$Rw & Pr$\to$Ar & Pr$\to$Cl & Pr$\to$Rw & Rw$\to$Ar & Rw$\to$Cl & Rw$\to$Pr & AVG \\ 
        \midrule
        Source only  & 34.9 & 50.0 & 58.0 & 37.4 & 41.9 & 46.2 & 38.5 & 31.2 & 60.4 & 53.9 & 41.2 & 59.9 & 46.1 \\
        JAN~\cite{long2017JAN} & 45.9 & 61.2 & 68.9 & 50.4 & 59.7 & 61.0 & 45.8 & 43.4 & 70.3 & 63.9 & 52.4 & 76.8 & 58.3 \\
        DWT~\cite{roy2019DWT} & 50.3 & 72.1 & 77.0 & 59.2 & 69.3 & 70.2 & 58.3 & 48.1 & 77.3 & 69.3 & 53.6 & 82.0 & 65.6 \\
        BSP~\cite{chen2019BSP} & 52.0 & 68.6 & 76.1 & 58.0 & 70.3 & 70.2 & 58.6 & 50.2 & 77.6 & 72.2 & 59.3 & 81.9 & 66.3 \\
        AFN~\cite{xu2019AFN} & 52.0 & 71.7 & 76.3 & 64.2 & 69.9 & 71.9 & 63.7 & 51.4 & 77.1 & 70.9 & 57.1 & 81.5 & 67.3 \\
        MDD~\cite{zhang2019MDD} & 54.9 & 73.7 & 77.8 & 60.0 & 71.4 & 71.8 & 61.2 & 53.6 & 78.1 & 72.5 & 60.2 & 82.3 & 68.1 \\
        STAFF~\cite{chen2020structure} & 53.3 & 71.9 & \textbf{80.2} & 63.1 & 69.8 & 74.1 & 65.3 & 50.9 & 77.8 & 73.1 & 56.6 & 82.4 & 68.2 \\
        CDAN-GD~\cite{cui2020gvb} & 55.3 & 74.1 & 78.2 & 62.4 & 72.6 & 71.8 & 63.8 & 54.1 & 80.1 & 73.1 & 58.7 & 83.6 & 69.0 \\
        DANCE~\cite{saito2020DANCE} & 54.3 & \textbf{75.9} & 78.4 & 64.8 & 72.1 & 73.4 & 63.2 & 53.0 & 79.4 & 73.0 & 58.2 & 82.9 & 69.1 \\
        \midrule
        DANN~\cite{ganin2015DANN} & 45.6 & 59.3 & 70.1 & 47.0 & 58.5 & 60.9 & 46.1 & 43.7 & 68.5 & 63.2 & 51.8 & 76.8 & 57.6 \\
        DANN+RN & 47.3 & 63.1 & 74.4 & 57.1 & 64.7 & 68.4 & 55.2 & 47.8 & 75.9 & 68.9 & 53.5 & 79.3 &  63.0\\
        \midrule
        CDAN~\cite{long2018CDAN}& 50.7 & 70.6 & 76.0 & 57.6 & 70.0 & 70.0 & 57.4 & 50.9 & 77.3 & 70.9 & 56.7 & 81.6 & 65.8 \\
        CDAN+RN & \textbf{55.6} & 72.6 & 78.1 & \textbf{65.7} & \textbf{74.7} & \textbf{74.6} & \textbf{66.2} & \textbf{57.1} & \textbf{82.0} & \textbf{75.2} & \textbf{60.5} & \textbf{84.6} & \textbf{70.6} \\
        \midrule
        \midrule
        \textbf{PDA} & Ar$\to$Cl & Ar$\to$Pr & Ar$\to$Rw & Cl$\to$Ar & Cl$\to$Pr & Cl$\to$Rw & Pr$\to$Ar & Pr$\to$Cl & Pr$\to$Rw & Rw$\to$Ar & Rw$\to$Cl & Rw$\to$Pr & AVG \\
        \midrule
        Source only & 46.3 & 67.5 & 75.9 & 59.1 & 59.9 & 62.7 & 58.2 & 41.8 & 74.9 & 67.4 & 48.2 & 74.2 & 61.3 \\
        DANN~\cite{ganin2015DANN} & 35.5 & 48.2 & 51.6 & 35.2 & 35.4 & 41.4 & 34.8 & 31.7 & 46.2 & 47.5 & 34.7 & 49.0 & 40.9 \\
        IWAN~\cite{zhang2018IWAN} & 53.9 & 54.5 & 78.1 & 61.3 & 48.0 & 63.3 & 54.2 & 52.0 & 81.3 & 76.5 & 56.8 & 82.9 & 63.6 \\
        SAN~\cite{cao2018SAN} & 44.4 & 68.7 & 74.6 & 67.5 & 65.0 & 77.8 & 59.8 & 44.7 & 80.1 & 72.2 & 50.2 & 78.7 & 65.3 \\
        DANCE~\cite{saito2020DANCE} & 53.6 & 73.2 & 84.9 & 70.8 & 67.3 & 82.6 & 70.0 & 50.9 & 84.8 & 77.0 & 55.9 & 81.8 & 71.1 \\
        AFN~\cite{xu2019AFN} & 58.9 & 76.3 & 81.4 & 70.4 & 73.0 & 77.8 & 72.4 & 55.3 & 80.4 & 75.8 & 60.4 & 79.9 & 71.8 \\
        JUMBOT~\cite{fatras2021JUMBOT} & 62.7 & 77.5 & 84.4 & \textbf{76.0} & 73.3 & 80.5 & 74.7 & 60.8 & 85.1 & 80.2 & \textbf{66.5} & 83.9 & 75.5 \\
        \midrule
        ETN~\cite{cao2019ETN}& 52.9 & 78.2 & 83.2 & 70.2 & 69.4 & 77.6 & 69.5 & 50.8 & 81.0 & 76.3 & 54.5 & 82.0 & 70.5 \\
        ETN+RN & 56.1 & 79.5 & 87.2 & 74.8 & 68.2 & 79.4 & 77.0 & 52.2 & 83.9 & \textbf{82.2} & 58.7 & 83.5 & 73.6 \\
        \midrule
        BA$^{3}$US~\cite{liang2020BA3US} & {60.6} & \textbf{83.2} & \textbf{88.4} & 71.8 & 72.8 & \textbf{83.4} & 75.5 & {61.6} & {86.5} & 79.3 & {62.8} & {86.1} & {76.0} \\
        BA$^{3}$US+RN & \textbf{63.5} & \textbf{83.2} & 88.3 & 72.8 & \textbf{73.4} & \textbf{83.4} & \textbf{77.2} & \textbf{62.6} & \textbf{87.7} & 80.8 & 63.6 & \textbf{87.0} & \textbf{77.0} \\
        \bottomrule
    \end{tabular}
    } 
    } 
    \vspace{-3mm}
\end{table*}

\subsubsection{Baselines}
Besides compared with existing normalization modules for domain adaptation (\ie, BN~\cite{ioffe2015BN}, AutoDIAL~\cite{cariucci2017autodial}, DSBN~\cite{chang2019domain}, and TN~\cite{wang2019TN}), we select popular \SOTA~ approaches as the baselines in three typical scenarios:
\begin{itemize}
    \item On the vanilla closed-set \textit{UDA}, we compare with DAN~\cite{long2015DAN}, DANN~\cite{ganin2015DANN}, JAN~\cite{long2017JAN}, MCD~\cite{saito2018MCD}, 
    CDAN~\cite{long2018CDAN}, iCAN~\cite{zhang2018iCAN}, DTA~\cite{Lee2019DTA}, DWT~\cite{roy2019DWT}, BSP~\cite{chen2019BSP}, AFN~\cite{xu2019AFN}, 
    CRST~\cite{zou2019CRST}, CADA~\cite{kurmi2019CADA}, MDD~\cite{zhang2019MDD}, CAN~\cite{kang2019CAN}, BNM~\cite{cui2020BNM}, DCAN~\cite{li2020DCAN}, IAA~\cite{jiang2020implicit}, STAFF~\cite{chen2020structure}, 
    GVB on CDAN (CDAN-GD)~\cite{cui2020gvb}, DMRL~\cite{wu2020dual}, DANCE~\cite{saito2020DANCE}, and DWL~\cite{xiao2021DWL}.
    
    \item On the \textit{PDA}, we compare with DANN~\cite{ganin2015DANN}, IWAN~\cite{zhang2018IWAN}, SAN~\cite{cao2018SAN}, AFN~\cite{xu2019AFN}, ETN~\cite{cao2019ETN}, BA$^3$US~\cite{liang2020BA3US}, DANCE~\cite{saito2020DANCE}, and JUMBOT~\cite{fatras2021JUMBOT}.
 
    \item On the \textit{MSDA}, we compare with DANN~\cite{ganin2015DANN}, D-CORAL~\cite{sun2016DCORAL}, CDAN~\cite{long2018CDAN}, Meta-MCD~\cite{li2020online} and SImpAl~\cite{venkat2020SImpAl} (two recently proposed MSDA methods).
\end{itemize}

For fair comparison, we run the proposed method three times with different random seeds and record the average results.
For clear comparison, we also verify the superiority of the proposed RN over normalization counterparts: BN~\cite{ioffe2015BN}, AutoDIAL~\cite{cariucci2017autodial}, DSBN~\cite{chang2019domain}, and TN~\cite{wang2019TN} (please refer to Table~\ref{tab:Regularizer_ThreeDatasets}).

\begin{table}
    \centering
    \caption{Accuracy ($\%$) on ImageCLEF-DA benchmark for ResNet-50 based UDA. The best results are in \textbf{bold}.}
    \label{tab:ImageCLEFDA_results}
    \small{
    \setlength{\tabcolsep}{1.5mm}{
    \begin{tabular}{lccccccc}
        \toprule
        Method & I$\to$P & P$\to$I & I$\to$C & C$\to$I & C$\to$P & P$\to$C & AVG \\
        \midrule
        Source only & 74.8 & 83.9 & 91.5 & 78.0 & 65.5 & 91.2 & 80.7 \\
        DAN~\cite{long2015DAN} & 74.5 & 82.2 & 92.8 & 86.3 & 69.2 & 89.8 & 82.5 \\
        DANN~\cite{ganin2015DANN} & 75.0 & 86.0 & 96.2 & 87.0 & 74.3 & 91.5 & 85.0 \\
        JAN~\cite{long2017JAN} & 76.8 & 88.0 & 94.7 & 89.5 & 74.2 & 91.7 & 85.8 \\
        iCAN~\cite{zhang2018iCAN} & 79.5 & 89.7 & 94.7 & 89.9 & 78.5 & 92.0 & 87.4 \\
        CAN~\cite{kang2019CAN} & 77.2 & 90.3 & 96.0 & 90.9 & 78.0 & \textbf{95.6} & 88.0 \\
        DMRL~\cite{wu2020dual} & 77.3 & 90.7 & 97.4 & 91.8 & 76.0 & 94.8 & 88.0 \\
        CADA~\cite{kurmi2019CADA} & 78.0 & 90.5 & 96.7 & 92.0 & 77.2 & 95.5 & 88.3 \\
        DCAN~\cite{li2020DCAN} & \textbf{80.5} & 91.2 & 95.7 & 91.8 & 77.2 & 93.3 & 88.3 \\
        \midrule
        DANN~\cite{ganin2015DANN} & 75.0 & 86.0 & 96.2 & 87.0 & 74.3 & 91.5 & 85.0 \\
        DANN+RN & 78.1 & 90.1 & 96.3 & 91.7 & 78.0 & 94.0 & 88.0\\
        \midrule
        CDAN~\cite{long2018CDAN} & 77.7 & 90.7 & \textbf{97.7} & 91.3 & 74.2 & 94.3 & 87.7 \\
        CDAN+RN & 78.6 & \textbf{92.7} & 97.2 & \textbf{92.8} & \textbf{79.1} & 94.8 & \textbf{89.2} \\
        \bottomrule
    \end{tabular}
    }
    }
    \vspace{-4mm}
\end{table}

\subsubsection{Implementation Details}
Without loss of generality, we adopt four popular domain adaptation methods as the test-beds: DANN~\cite{ganin2015DANN}, CDAN~\cite{long2018CDAN}, ETN~\cite{cao2019ETN},
and BA$^3$US~\cite{liang2020BA3US}.
On one backbone network (\eg, ResNet-50~\cite{he2016resnet}) pretrained on ImageNet, we replace all the BN~\cite{ioffe2015BN} within different intermediate layers in the backbone with our RN without changing the original settings.
We initialize the parameters of RA to unit vectors and constrain their weights to be in the range $[0.5,1]$.
It should be pointed out that the substitution works without an additional pre-training procedure on ImageNet dataset, and it is flexible for practical usage. The flexible replacement indicates the versatility of our RN. 
We implement the RN via PyTorch~\cite{paszke2019pytorch}.
For fair comparison, the training configurations (\eg, data pipeline, batch-size, learning rate, 
optimization algorithm) are all the same as the original baselines except the normalization module which are replaced by our RN.
We conduct the experiments of RN with $3$ random seeds and report the average accuracies.

\subsection{Evaluation Results}
\subsubsection{Results on Small-scale Dataset}
First, we conduct the comparison experiments on ImageCLEF-DA, one popular small-scale cross-domain benchmark.
We adopt ResNet-50 as the backbone network and choose DANN and CDAN as the test-bed methods.
As listed in Table~\ref{tab:ImageCLEFDA_results}, on the average performance of 6 adaptation scenarios, the proposed RN helps DANN and CDAN promote their classification accuracies by $3.0\%$ and $1.5\%$, respectively. The results demonstrate the effectiveness of our RN. For the comparisons of RN and existing normalization modules on ImageCLEF-DA, please refer to Table~\ref{tab:Regularizer_ThreeDatasets}.

\begin{table}
    \centering
    \caption{The average classification accuracies over $12$ classes ($\%$) on VisDA-C for vanilla closed-set UDA on both ResNet-50 and ResNet-101. The best results are in \textbf{bold}.}
    \label{tab:ResultsOnVisDA2017}
    \small{
    \setlength{\tabcolsep}{2.4mm}{
    \begin{tabular}{lcc}
        \toprule
        \multirow{1}{*}{Method} 
         & ResNet-50 & ResNet-101 \\
        \midrule
        Source only & - & 52.4 \\
        JAN~\cite{long2017JAN} & 61.6 & - \\
        DAN~\cite{long2015DAN} & 61.6 & 62.8 \\
        MCD~\cite{saito2018MCD} & 69.7 & 71.9 \\
        DMRL~\cite{wu2020dual} & - & 75.5 \\
        IAA~\cite{jiang2020implicit} & 75.8 & - \\
        BSP~\cite{chen2019BSP} & - & 75.9 \\
        AFN~\cite{xu2019AFN} & - & 76.1 \\
        DWL & - & 77.1 \\
        CRST~\cite{zou2019CRST} & - & 77.9 \\
        DANCE~\cite{saito2020DANCE} & 70.2 & - \\
        JUMBOT~\cite{fatras2021JUMBOT} & 72.5 &  \\
        CDAN-GD~\cite{cui2020gvb} & 74.9 & - \\
        DTA~\cite{Lee2019DTA} & 76.2 & - \\
        \midrule
        DANN~\cite{ganin2015DANN} & 54.9 & 57.4 \\
        DANN+RN & 71.4 & 74.9 \\
        \midrule
        CDAN~\cite{long2018CDAN} & 70.0 & 73.9 \\
        CDAN+RN & \textbf{79.6} & \textbf{80.1} \\
        \bottomrule
    \end{tabular}
    }
    }
    \vspace{-4mm}
\end{table}

\begin{table*}
    \centering
    \caption{Classification accuracies ($\%$) of different normalization methods on three domain adaptation benchmarks for UDA.}
    \label{tab:Regularizer_ThreeDatasets}
    \small{
    \setlength{\tabcolsep}{0.8mm}{
    \begin{tabular}{l|ccccccc|cc|ccccc}
        \toprule
        \multirow{2}{*}{Method} & \multicolumn{7}{c|}{ImageCLEF-DA} & \multicolumn{2}{c|}{VisDA-C} & \multicolumn{5}{c}{Office-Home} \\
         & I$\to$P & P$\to$I & I$\to$C & C$\to$I & C$\to$P & P$\to$C & AVG & ResNet-50 & ResNet-101 & Ar $\to$X & Cl $\to$X & Pr $\to$X & Rw $\to$X & AVG \\
        \midrule
        DANN(+BN)~\cite{ganin2015DANN} & 75.0 & 86.0 & 96.2 & 87.0 & 74.3 & 91.5 & 85.0 & 54.9 & 57.4 & 58.3 & 55.5 & 52.8 & 63.9 & 57.6 \\
        DANN+AutoDIAL~\cite{cariucci2017autodial} & 77.3 & 88.8 & 95.3 & 89.5 & \textbf{79.0} & 91.3 & 86.9 & 62.5 & 64.7 & 61.4 & 55.4 & 54.6 & 63.9 & 58.8 \\
        DANN+DSBN~\cite{chang2019domain} & 77.2 & 88.2 & 93.8 & 90.3 & 77.8 & 89.3 & 86.1 & 65.0 & 69.6 & 57.0 & 55.2 & 50.2 & 56.8 & 54.8\\
        DANN+TN~\cite{wang2019TN} & \textbf{78.2} & 89.5 & 95.5 & 91.0 & 76.0 & 91.5 & 87.0 & 66.3 & - & 58.8 & 58.3 & 55.6 & 64.6 & 59.3 \\
        DANN+RN & 78.1 & \textbf{90.1} & \textbf{96.3} & \textbf{91.7} & 78.0 & \textbf{94.0} & \textbf{88.0} & \textbf{71.4} & \textbf{74.9} & \textbf{61.6} & \textbf{63.4} & \textbf{59.6} & \textbf{67.2} & \textbf{63.0} \\ 
        \midrule
        CDAN(+BN)~\cite{long2018CDAN} & 77.7 & 90.7 & \textbf{97.7} & 91.3 & 74.2 & 94.3 & 87.7 & 70.0 & 73.9 & 65.8 & 65.9 & 61.9 & 69.7 & 65.8 \\
        CDAN+AutoDIAL~\cite{cariucci2017autodial} & 77.8 & 90.3 & 96.8 & 91.2 & 77.2 & 94.5 & 88.0 & 71.8 & 74.5 & 65.3 & 66.4 & 61.8 & 73.9 & 67.4 \\
        CDAN+DSBN~\cite{chang2019domain} & 76.2 & 92.2 & 94.9 & 90.1 & 74.0 & 94.3 & 86.9 & 72.9 & 78.6 & 65.5 & 65.0 & 58.1 & 66.7 & 64.1 \\
        CDAN+TN~\cite{wang2019TN} & 78.3 & 90.8 & 96.7 & 92.3 & 78.0 & 94.8 & 88.5 & 71.4 & - & 66.3 & 68.4 & 64.5 & 71.3 & 67.6 \\
        CDAN+RN & \textbf{78.6} & \textbf{92.7} & 97.2 & \textbf{92.8} & \textbf{79.1} & \textbf{94.8} & \textbf{89.2} & \textbf{79.6} & \textbf{80.1} & \textbf{68.8} & \textbf{71.7} & \textbf{68.4} & \textbf{73.4} & \textbf{70.6} \\
        \bottomrule
    \end{tabular}
    }
    }
\end{table*}

\subsubsection{Results on Medium-scale Dataset}
Next, we summarize the results of UDA and PDA experiments on Office-Home in Table~\ref{tab:OfficeHome_results}. For fair comparison in PDA scenario, we follow the protocol of ETN~\cite{cao2019ETN} and BA$^3$US~\cite{liang2020BA3US}~\footnote{\url{https://github.com/tim-learn/BA3US}.} in the PDA experiments.
Experimental results show that the proposed RN promotes the performance of CDAN by $4.8\%$ in UDA. In PDA, our RN also benefits ETN by $3.1\%$ and BA$^3$US by $1.0\%$.
It is also noteworthy that 
in PDA experiments on Office-Home, there is a large semantic difference between the two domains: source domain contains $65$ classes while target domain contains only $25$ classes.
Despite the differences, our RN still helps ETN achieve better performance in average accuracy (from $70.5\%$ to $73.6\%$).
Consequently, these numerical results ensure the versatility of the proposed RN in DA.
It is convincing that RN can consistently helps boost performance of UDA methods.
For the visualization of learned visual feature (\eg, tSNE~\cite{tsne}), please refer to Section~\ref{sup_sec:Visualization}.

\subsubsection{Results on Large-scale Dataset}
To further demonstrate the effectiveness of the proposed RN, we conduct evaluation experiments on VisDA-C benchmark.
We compare several popular UDA methods and evaluate their classification accuracies in Table~\ref{tab:ResultsOnVisDA2017}.
We observe that our RN helps CDAN achieve better performance on ResNet-50 (by $9.6\%$) and ResNet-101 (by $6.2\%$).
Additionally, the gap between CDAN+RN on ResNet-50 and that on ResNet-101 is only 0.5\%, indicating our RN is particularly effective to conduct domain adaptation in the large-scale dataset.
The possible reason is the domain statistics would be more accurate when the dataset is large, which is more beneficial to our RN.


\subsubsection{Different Normalization and General Regularizer}
To verify that our RN can work as a general regularizer in domain adaptation scenarios, without loss of generality, we choose DANN~\cite{ganin2015DANN} and CDAN~\cite{long2018CDAN} as the test-bed methods and conduct evaluation experiments on three benchmarks: Office-Home, ImageCLEF-DA, and VisDA-C.
For fair comparisons, existing $\SOTA$ feature normalization modules are considered: BN~\cite{ioffe2015BN}, AutoDIAL~\cite{cariucci2017autodial}, DSBN~\cite{chang2019domain}, and TN~\cite{wang2019TN}.
Noting that all following experiments of \textit{Method+DSBN} are conducted without extra constraints, \eg, pseudo labels algorithm or other loss functions.
Numerical results are listed in Table~\ref{tab:Regularizer_ThreeDatasets}, where ``Y$\to$X'' means that the domain ``Y'' respectively adapts to other three domains of Office-Home and we report the average accuracy of the three transfer tasks. 

As we observe in Table~\ref{tab:Regularizer_ThreeDatasets} that, 
our RN consistently offers larger improvements than other counterparts to a variety of domain adaptation methods on various datasets.
It affirms the effectiveness of the proposed RN beyond existing normalization counterparts in DA.
It is worth noting that DSBN~\cite{chang2019domain} even produces worse performance on small- and medium-scale datasets, indicating that separating $\gamma$ and $\beta$ may suffers from \textit{negative transfer} without the extra pseudo labels algorithms to conduct the target parameters update properly.
In addition, when it comes to computation cost, our RN achieves better trade-off between the time costs of train and test phases simultaneously.


\begin{table}
    \centering
    \caption{Accuracy (\%) on Office-Home benchmark for ResNet-50 based multi-source UDA. The best results are in \textbf{blod}.}
    \label{tab:OfficeHome_Results_MSDA}
    \small{
    \setlength{\tabcolsep}{1.8mm}{
    \begin{tabular}{lccccc}
        \toprule
        \textbf{Single Best} & $\to$Ar & $\to$Cl & $\to$Pr & $\to$Rw & AVG \\
        \midrule
        ResNet-50~\cite{he2016resnet} & 53.9 & 41.2 & 59.9 & 60.4 & 53.9 \\
        D-CORAL~\cite{sun2016DCORAL} & 67.0 & 53.6 & 80.3 & 76.3 & 69.3 \\
        DAN~\cite{long2015DAN} & 67.9 & 55.9 & 80.4 & 75.8 & 70.0 \\
        RevGrad~\cite{ganin2015DANN} & 67.9 & 55.9 & 80.4 & 75.8 & 70.0 \\
        \midrule
        CDAN(+BN)~\cite{long2018CDAN} & 70.9 & 56.7 & 81.6 & 77.3 & 71.6 \\
        CDAN+AutoDIAL~\cite{cariucci2017autodial} & 71.2 & 57.5 & 81.4 & 76.2 & 71.6 \\
        CDAN+DSBN~\cite{chang2019domain} & 70.2 &   51.4 & 78.4 & 78.4 & 69.6 \\
        CDAN+TN~\cite{wang2019TN} & 71.9 & 59.0 & 82.9 &  79.5 & 73.3 \\
        CDAN+RN & \textbf{75.2} & \textbf{60.5} & \textbf{84.6} & \textbf{82.0} & \textbf{75.6} \\
        \midrule
        \textbf{Combination} & $\to$A & $\to$Cl & $\to$Pr & $\to$Rw & AVG \\
        \midrule
        ResNet-50~\cite{he2016resnet} & 65.3 & 49.6 & 79.7 & 75.4 & 67.5 \\
        D-CORAL~\cite{sun2016DCORAL} & 68.1 & 58.6 & 79.5 & 82.7 & 72.2 \\
        DAN~\cite{long2015DAN} & 68.4 & 59.1 & 79.5 & 82.7 & 72.4 \\
        RevGrad~\cite{ganin2015DANN} & 68.4 & 59.1 & 79.5 & 82.7 & 72.4\\
        Meta-MCD~\cite{li2020online} & 70.2 & 60.5 & 81.2 & 83.1 & 73.8 \\
        SImpAl~\cite{venkat2020SImpAl} & 73.4 & 62.4 & 81.0 & 82.7 & 74.8 \\
        \midrule
        CDAN(+BN)~\cite{long2018CDAN} & 71.4 & 64.2 & 81.1 & 82.3 & 74.8 \\
        CDAN+AutoDIAL~\cite{cariucci2017autodial} & 75.7 & 64.2 & 83.7 & 83.9 & 76.9 \\
        CDAN+DSBN~\cite{chang2019domain} & 71.7 & 57.2 & 77.5 & 79.1 & 71.4 \\
        CDAN+TN~\cite{wang2019TN} & 74.7 & 64.6 & 83.1 & 83.3 & 76.4 \\
        CDAN+RN & \textbf{75.6} & \textbf{66.8} & \textbf{85.3} & \textbf{85.3} & \textbf{78.3} \\
        \bottomrule
    \end{tabular}
    }
    }
\end{table}

\subsubsection{Results on MSDA}
To demonstrate the versatility of our RN, we also conduct multi-source domain adaptation (MSDA) on Office-Home benchmark.
We choose CDAN~\cite{long2018CDAN} as the baseline and also compare with existing normalization counterparts. 
The results are listed in 
Table~\ref{tab:OfficeHome_Results_MSDA}.
For simplicity: ``$\to$X'' denotes the adaptation task from other three domains to ``X'' domain.
\textit{Single Best} denotes the best performance of all tasks on single-source domain adaptation, and \textit{Combination} refers to merging data from multiple source domains and constructing a new and larger source domain dataset.
Obviously, our RN models consistently outperform the other BN-variant modules for all settings.
On the AVG, the proposed RN module promotes CDAN by $4.0\%$ in \textit{Single Best} scenario and $3.5\%$ in \textit{Combination} scenario.
The improvements are higher than that from the existing normalization methods and CDAN+RN outperforms D-CORAL~\cite{sun2016DCORAL}, Meta-MCD~\cite{li2020online}, and SImpAl~\cite{venkat2020SImpAl}.
Consequently, the results indicate that RN is also versatile to benefit multi-source domain adaptation scenarios.

\begin{table}
    \centering
    \caption{Comparisons of training and testing time (s) of different feature normalization methods on the same domain adaptation baseline (CDAN + ResNet-50).
    }
    \label{spptab:test_time} 
    \small{
    \setlength{\tabcolsep}{1.8mm}{
    \begin{tabular}{lccccc}
        \toprule
        Method & \tabincell{c}{CDAN \\ (+BN)} & \tabincell{c}{CDAN \\ +AutoDIAL} & \tabincell{c}{CDAN \\ +TN} & \tabincell{c}{CDAN \\ +DSBN} & \tabincell{c}{CDAN \\ +Ours} \\
        \midrule
        Train & 0.13 & 0.38 & 0.71 & 0.14 & 0.50 \\
        \midrule
        Test & 1.91 & 3.76 & 64.40 & 5.17 & 1.90 \\
        \bottomrule
    \end{tabular}
    }
    }
\end{table}

\subsubsection{Train and test time comparison}
Besides, we also compare our RN with the other normalization modules in the perspective of computation cost, \ie, the time in both training and inference stages.
In specific, we report the quantitative values of computation cost of different normalization methods on the UDA task of Pr $\to$ Rw (Office-Home) with $4$ threads and one Tesla V100 GPU. The backbones are the ResNet-50 by default.
To eliminate the noise in the estimations, the training time is calculated based on the average of the time cost of $10,000$ iterations, including forward and backward operations.
And th test times are the total time cost of evaluating the whole target domain dataset.

These results are listed in the Table~\ref{spptab:test_time}.
We can observe that the training time cost of RN is more than vanilla BN, AutoDIAL, and DSBN, but less than TN.
On the other hand, the test time cost of RN is very close to the vanilla BN, and also less than other normalization methods, implying that our RN has more advantages in practical applications.
These results demonstrate that our RN achieves better trade-off between the time costs of both train and test phases and simultaneously achieves better performance.



\subsection{Ablation Study}
\label{Analysis}

\begin{table}[h]
    \centering
    \caption{Ablation study results of RC and RA-Gate with CDAN~\cite{long2018CDAN} (ResNet-50) on Office-Home benchmark.}
    \label{tab:ablation}
    \small{
    \setlength{\tabcolsep}{1.8mm}{
    \begin{tabular}{cc|ccccc}
        \toprule
        RC & RA & Ar $\to$ X & Cl $\to$ X & Pr $\to$ X & Rw $\to$ X & AVG \\
        \midrule
        - & - & 65.8 & 65.9 & 61.9 & 69.7 & 65.8\\
        \midrule
        -  & \checkmark & 67.3 & 69.4 & 65.5 & 72.5 & 68.9 \\
        \midrule
        \checkmark & \checkmark  & \textbf{68.8} & \textbf{71.7} & \textbf{68.4} & \textbf{73.4} & \textbf{70.6} \\
        \bottomrule
    \end{tabular}
    }
    }
\end{table}

\subsubsection{Ablation Study (RC \& RA)}
To investigate the effects of RC and RA, we conduct additional ablation study experiments on Office-Home as an instance.
The results are shown in Table~\ref{tab:ablation}, where ``$\to$X'' means to other $3$ domains and we report the average accuracy of the $3$ transfer tasks.
Based on CDAN+BN refer to the first row, we progressively add the RA and RC, respectively. Noting that RC cannot be trained independently.
It is clear that aggregating statistics of cross-domain corresponding channels (\ie, the second row) outperforms the baseline, and the full method (\ie, the third row) achieves the best performance.
It verifies the effectiveness of the exploitation of the correlation of cross-domain non-corresponding channels.
Similar observations can be found in other adaptation scenarios.
Therefore, indicating the effectiveness and necessity of the RC and RA in our RN.

\subsubsection{Influence of $g$ in RA}
To investigate the influence of $g$ in RA, we conduct quantitative analysis on the UDA task ``Art $\to$ Clipart'' from Office-Home dataset.

\begin{figure}
    \centering
    \includegraphics[width=\linewidth]{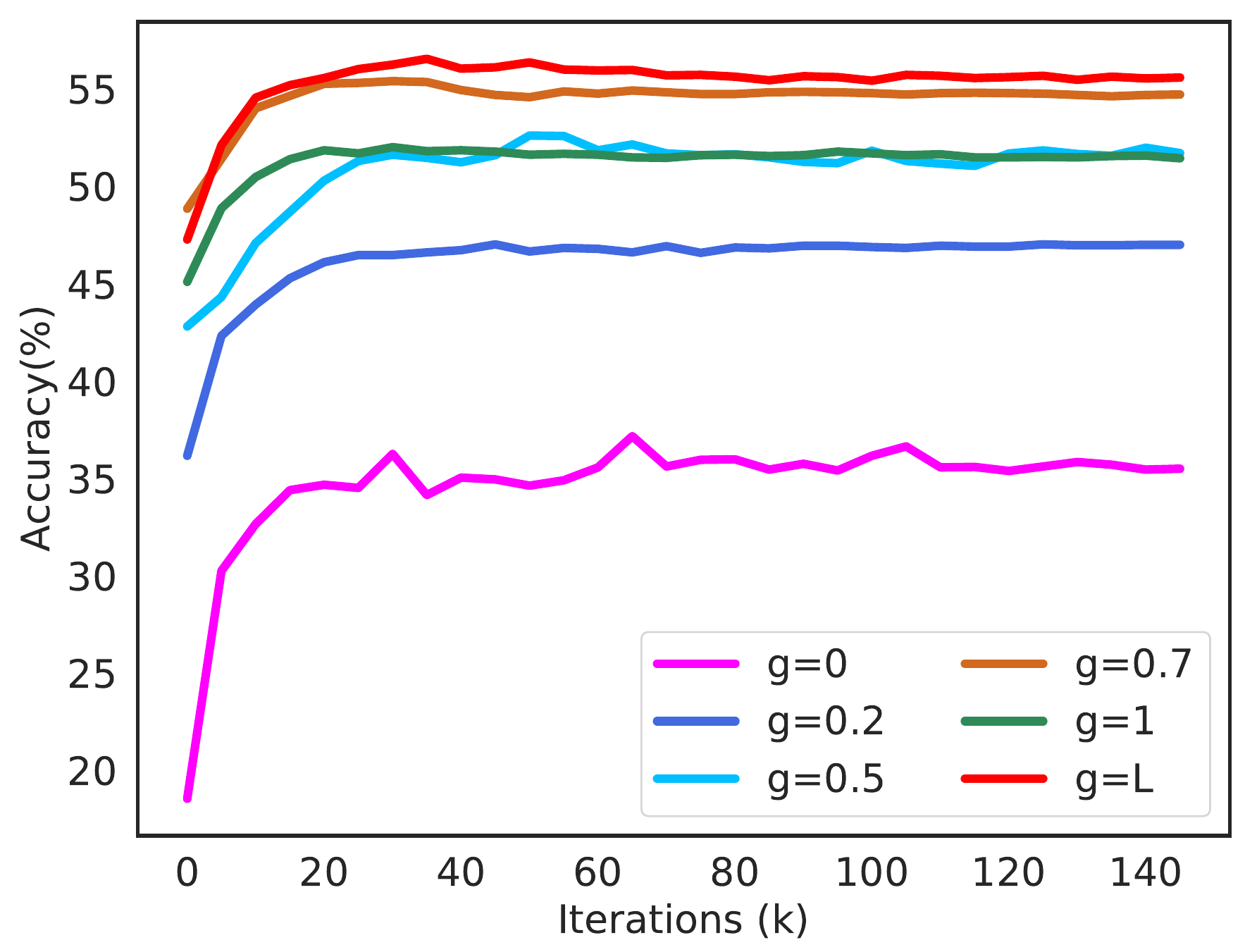}
    \caption{The influence of $g$ in RA. ``L'' means $g$ is learnable. The backbone is ResNet-50~\cite{he2016resnet} and the UDA method is CDAN~\cite{long2018CDAN}.}
    \label{analysis:RA-Gate}
\end{figure}

The results are visualized in Fig.~\ref{analysis:RA-Gate}. Noting that ``\textit{L}'' means $g$ is learnable.
Obviously, the learnable $g$ achieves the best performance.
The training curves demonstrate that the best value interval of RA should be 0.5$\sim$1, verifying that the effectiveness of our constraint on the learnable $g$.
It is easy to understand that $g=0$ and $g=1$ mean that training models only with outputs of RC and without outputs of RC, respectively, and $g < 0.5$ means that using less the original statistics leads to lose domain-specific information.
The experimental results 
also demonstrate the effectiveness of RC.
To sum up, both the RA and RC are effective and benefit to domain adaptation.

\begin{figure}
    \centering
    \includegraphics[width=\linewidth]{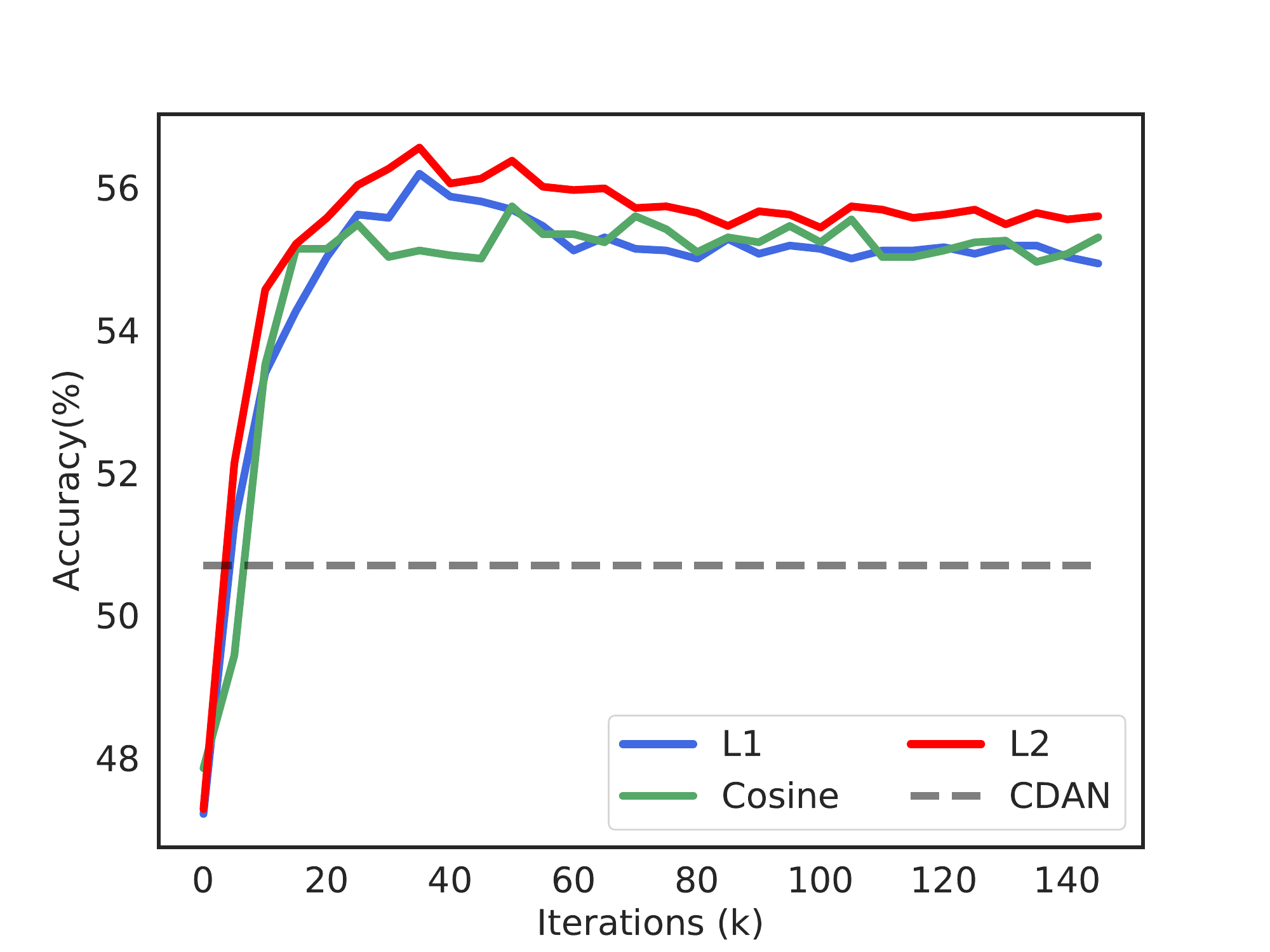}
    \caption{Classification accuracy of different correlation measures applied in RC. The dashed line indicates the performance of CDAN.}
    \label{analysis:distance}
    \vspace{-2mm}
\end{figure}

\subsection{Further Investigation}
\subsubsection{Analysis of Measures of Correlations}
To explain the importance of $l_2$ distance in calculating correlations, we compare the popular different distance metrics on correlations measures. 
The experiments are conducted on the UDA task ``Art$\to$Clipart'' in Office-Home dataset.

As illustrated in Fig.~\ref{analysis:distance}, the $l_2$ distance achieves the best performance because it enables the channels with similar patterns to have larger weights.
Additionally, the $l_2$ obtains the results with small margin ($< 1\%$) than other measures, indicating the RC is robust to different distance measures. 
Moreover, with different distance measures, our RN consistently obtains significant improvement over the baseline method, indicating the effectiveness of our RN.
Similar observations can also be found in other DA scenarios.

\begin{figure}
    \centering
    \includegraphics[width=0.88\linewidth]{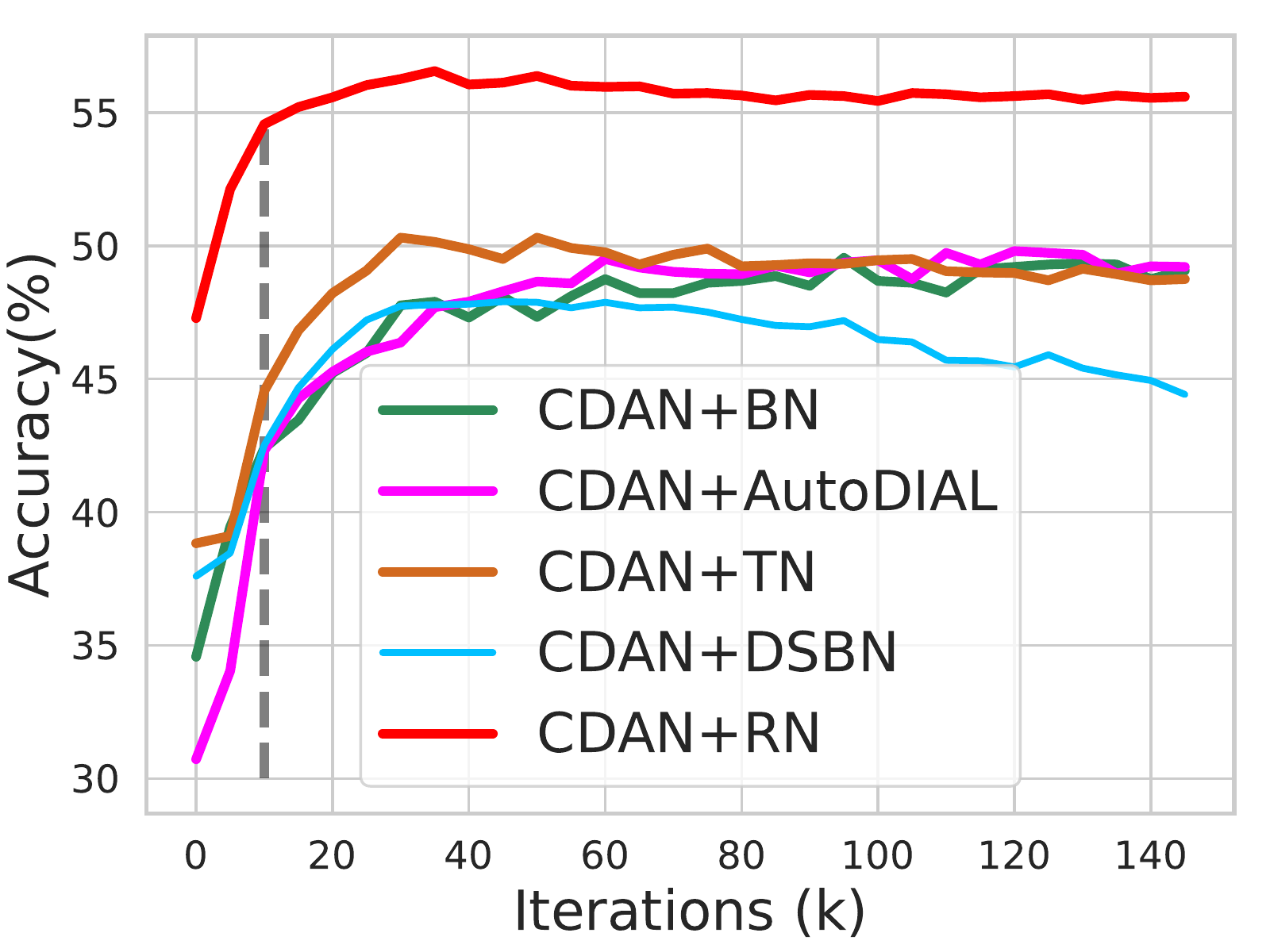}
    \caption{Training convergence analysis of various feature normalization techniques when the backbone is ResNet-50~\cite{he2016resnet} and the UDA baseline method is CDAN~\cite{long2018CDAN}.}
    \label{fig:TrainingConvergence}
    \vspace{-4mm}
\end{figure}

\begin{figure*}
    \centering
     \subfigure[A-distance]{
     \label{analysis:a_distance}
            \includegraphics[width=0.48\linewidth]{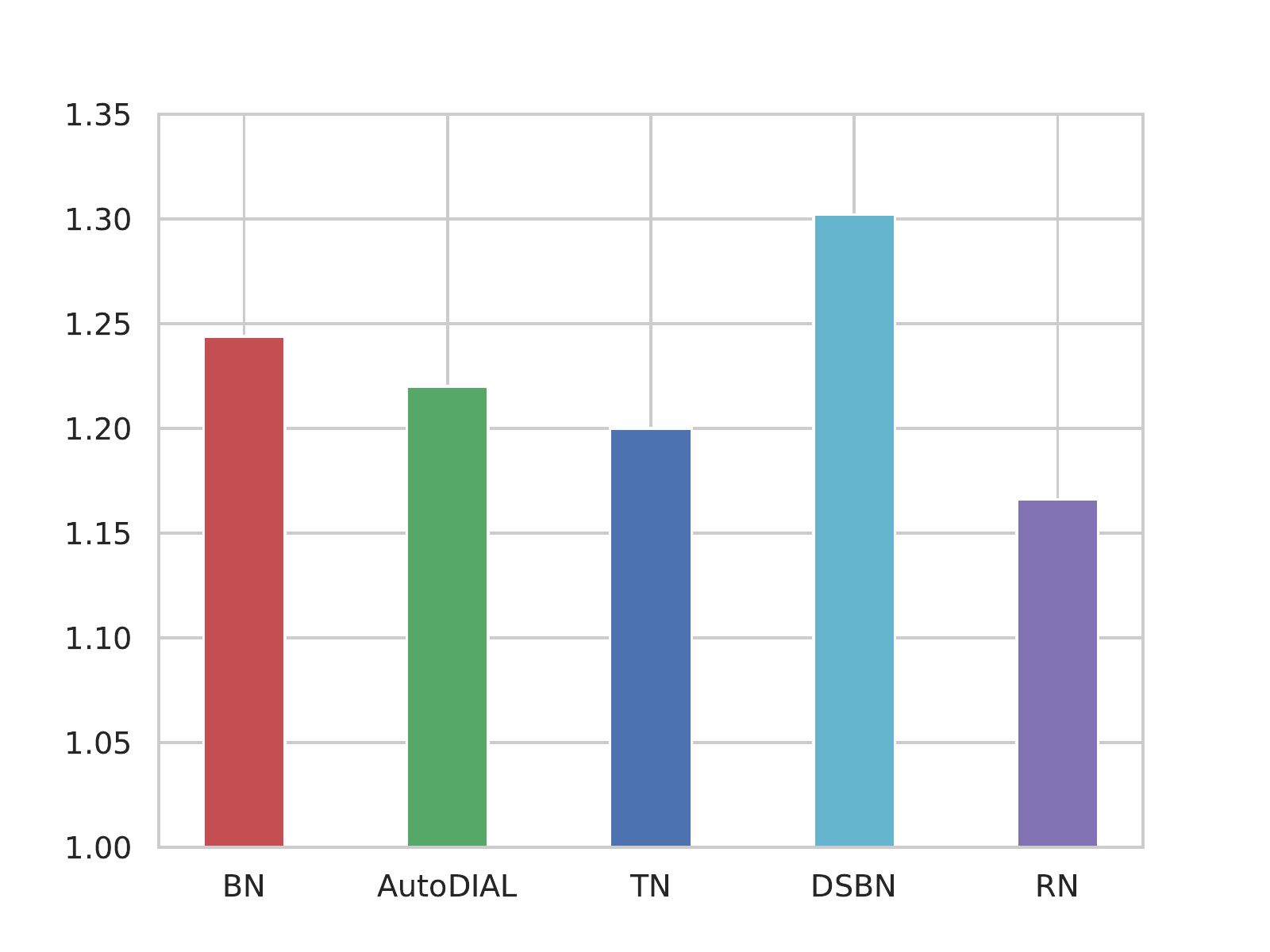}
         }
    \subfigure[$\lambda$]{
        \label{analysis:lambda}
            \includegraphics[width=0.48\linewidth]{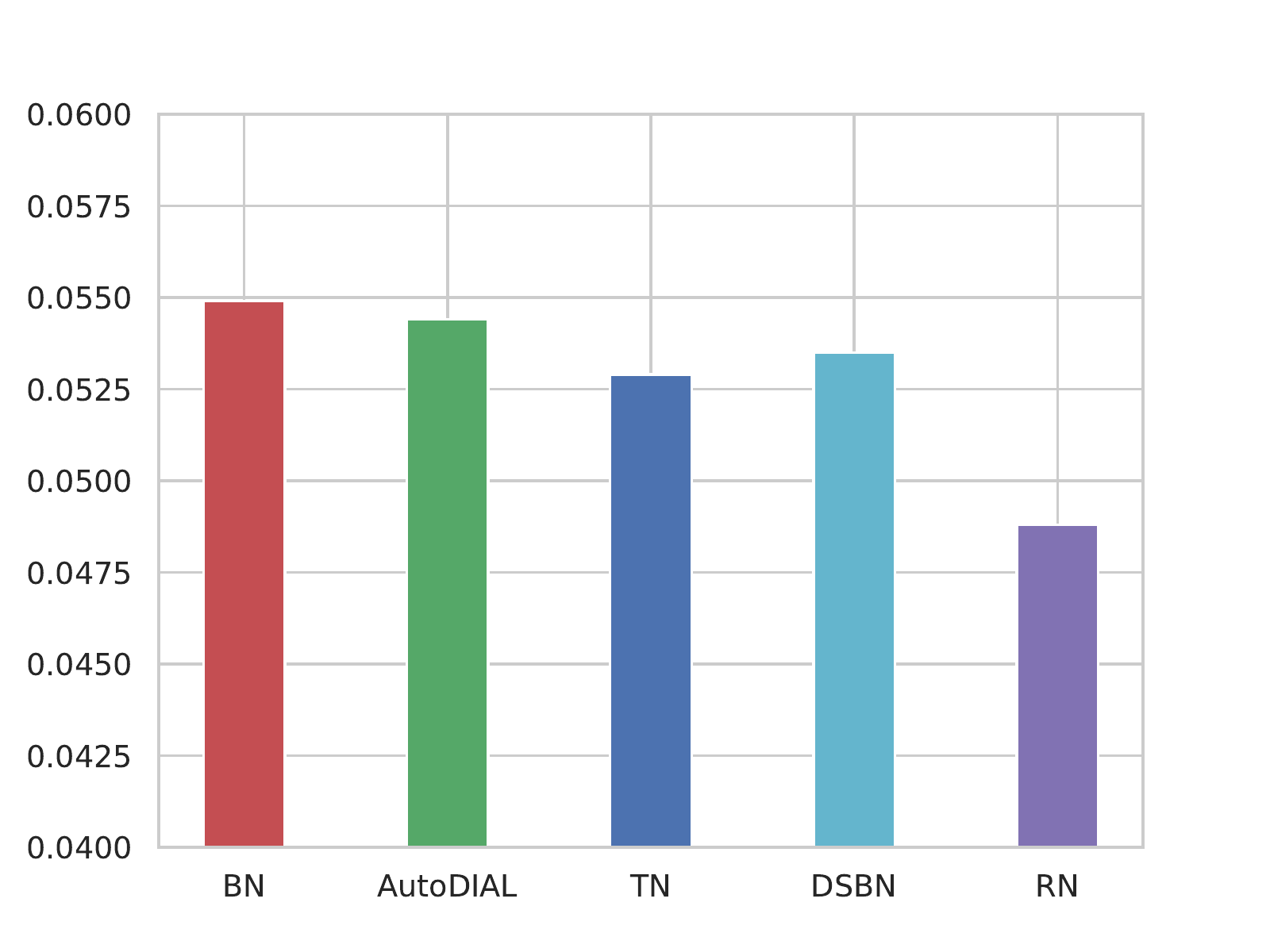}
         }
    \label{analysis:theoretical}
    \caption{
    Theoretical analysis of A-distance (a) and $\lambda$ in Eq.~\ref{eq:BenTheory} (b) when using different kinds of feature normalization methods on the same domain adaptation method (\eg, CDAN).}
    \label{fig:analysis}
\end{figure*}

\subsubsection{Training Convergence}
To illustrate the convergence performance and training stability of our RN, we present the classification accuracy during training on the UDA task Art $\to$ Clipart of Office-Home.
The similar training curves are observed in other adaptation scenarios.

As illustrated in Fig.~\ref{fig:TrainingConvergence}, the proposed RN fast and stably converges to the best accuracy, and achieves the optimal accuracy of over $95\%$ with only $1,000$ training iterations (black dotted line), compared with other existing normalization counterparts. We also notice that the accuracy curve of DSBN drops after $60$ training iterations, and it indicates that CDAN with DSBN suffers from negative transfer. This observation also verifies the importance of feature normalization module in domain adaptation tasks.

\subsection{Theoretical Understanding}
\label{sup_sec:theory}

\subsubsection{Theoretical Insight}
As Ben~\etal~\cite{ben2010theory} pioneered, 
the learning bound of domain adaptation is:
\begin{equation}
    \mathbf{\varepsilon}_{\Tdomain}(h) \leq \mathbf{\varepsilon}_{\Sdomain}(h) + \frac{1}{2} {d_{\mathcal{H} \Delta \mathcal{H}}(\Sdomain, \Tdomain)} + \lambda.
    \label{eq:BenTheory}
\end{equation}
Eq.~\ref{eq:BenTheory} bounds the expected risk $\mathbf{\varepsilon}_{\Tdomain}(h)$ of a hypothesis $h$ on the target domain by: 
1) the expected risk of $h$ on the source domain, $\mathbf{\varepsilon}_{\Sdomain}(h)$;
2) the A-distance~\cite{ben2010theory}, ${d_{\mathcal{H} \Delta \mathcal{H}}(\Sdomain, \Tdomain)} = 2(1-2\epsilon)$, a domain-divergence measure, where the $\epsilon$ is the error rate of a domain classifier which is trained to discriminate source and target domains;
3) the risk $\lambda$ of an ideal joint hypothesis $h^*$ for both source and target domains.
The A-distance and the $\lambda$ helps us better understand the rationale of one certain method in the topic of domain adaptation.

To further investigate the theoretical advantage of the proposed RN beyond the existing normalization modules, we estimate the A-distance and the $\lambda$ on the adaptation task of A $\to$ C (Office-Home dataset) with CDAN + various normalization methods. The results are shown in Fig.~\ref{analysis:a_distance} and Fig.~\ref{analysis:lambda} that, 
Compared with the other normalization counterparts, our RN helps CDAN obtain lower values in both A-distance and $\lambda$.
It indicates that the proposed RN facilitates more transferable representation from the perspectives of $d_{\mathcal{H} \Delta \mathcal{H}}(\Sdomain, \Tdomain)$ and $\lambda$.
And consequently, after learning visual representation with better transferability, the proposed RN is able to obtain better generalization performance.

\begin{table}[h]
    \centering
    \caption{To better understand the rationale of the propose method, we investigate the distance of the nearest channels across domains in the last feature normalization module in \texttt{stage4} on UDA task Art $\to$ Clipart and ResNet-50~\cite{he2016resnet} is adopted as the backbone. ``\%'' denotes the ratio of the corresponding channels in the nearest channels across domains.}
    \label{tab:channel_dist}
    \small{
    \setlength{\tabcolsep}{2.4mm}{
    \begin{tabular}{l|cccc|c}
        \toprule
        \multirow{2}{*}{Method} & \multicolumn{4}{c|}{\texttt{Stage}} & \multirow{2}{*}{$\sum$} \\
        \cline{2-5}
          & \texttt{1} & \texttt{2} & \texttt{3} & \texttt{4} &  \\
        \midrule
        \multirow{2}{*}{CDAN+AutoDIAL} & 3.77 & 3.48 & 1.84 & 0.97 & 10.1 \\
         & 5.1\% & 3.5\% & 2.0\% & 1.4\% & - \\
        \midrule
        \multirow{2}{*}{CDAN+DSBN} & 5.23 & 2.25 & 1.57 & 0.88 & 9.9 \\
         & 3.9\% & 3.7\% & 0.8\% & 0.3\% & - \\
        \midrule
        \multirow{2}{*}{CDAN+TN} & 4.01 & 3.43 & 1.87 & 0.72 & 10.0\\
         & 5.1\% & 3.1\% & 2.8\% & 0.8\% & - \\
        \midrule
        \multirow{2}{*}{CDAN+Ours} & 3.06 & 3.31 & 1.52 & 0.62 & 8.5\\
         & 3.1\% & 5.6\% & 3.0\% & 1.4\% & - \\
        \bottomrule
    \end{tabular}
    }
        
        
       
        
       
    }
\end{table}

\subsubsection{Distance of the Nearest Channels Across Domains}
In Table~\ref{tab:channel_dist}, we calculate the distance of any two channels across domains in the last feature normalization module in each stage. The four stages denote the four ``layer'' in ResNet-50 with channel numbers as 256, 512, 1024, and 2048, respectively.
The distance is calculated as follows:
\begin{equation}
    d^{(j)} = | \frac{\mu_s^{(j)}}{\sqrt{\sigma_s^{2(j)}}} - \frac{\mu_t^{(j)}}{\sqrt{\sigma_t^{2(j)}}}  |,    
\end{equation}
where $j$ denotes the $j$-th channel, which is introduced by Wang \etal~\cite{wang2019TN}.
Noting that the source and target domains share the same mean and variance in BN, and we do not calculate the distance across domains.
The goal of RN is to find each channel's compensatory information and then aggregate them.
The compensatory  consists of the information of both the corresponding and non-corresponding channels, where the nearest channels across domains has the largest correlation weight.
Hence, we report the sum of the distances between all pairs of the nearest channels.
The smaller distance means the greater ability to align both corresponding and non-corresponding channels to a certain extent.
We can observe that RN obtains the smallest value, implying the better performance of alignment of RN than other methods.
Besides, among the nearest channels, the proportion of corresponding channel is very small, verifying the misalignment between corresponding channels across domains.

\begin{figure*}
\subfigure[A$\to$C (CDAN)]{
  \centering
    $\underbrace{
    \begin{minipage}{0.18\linewidth}
        \includegraphics[width=1\linewidth]{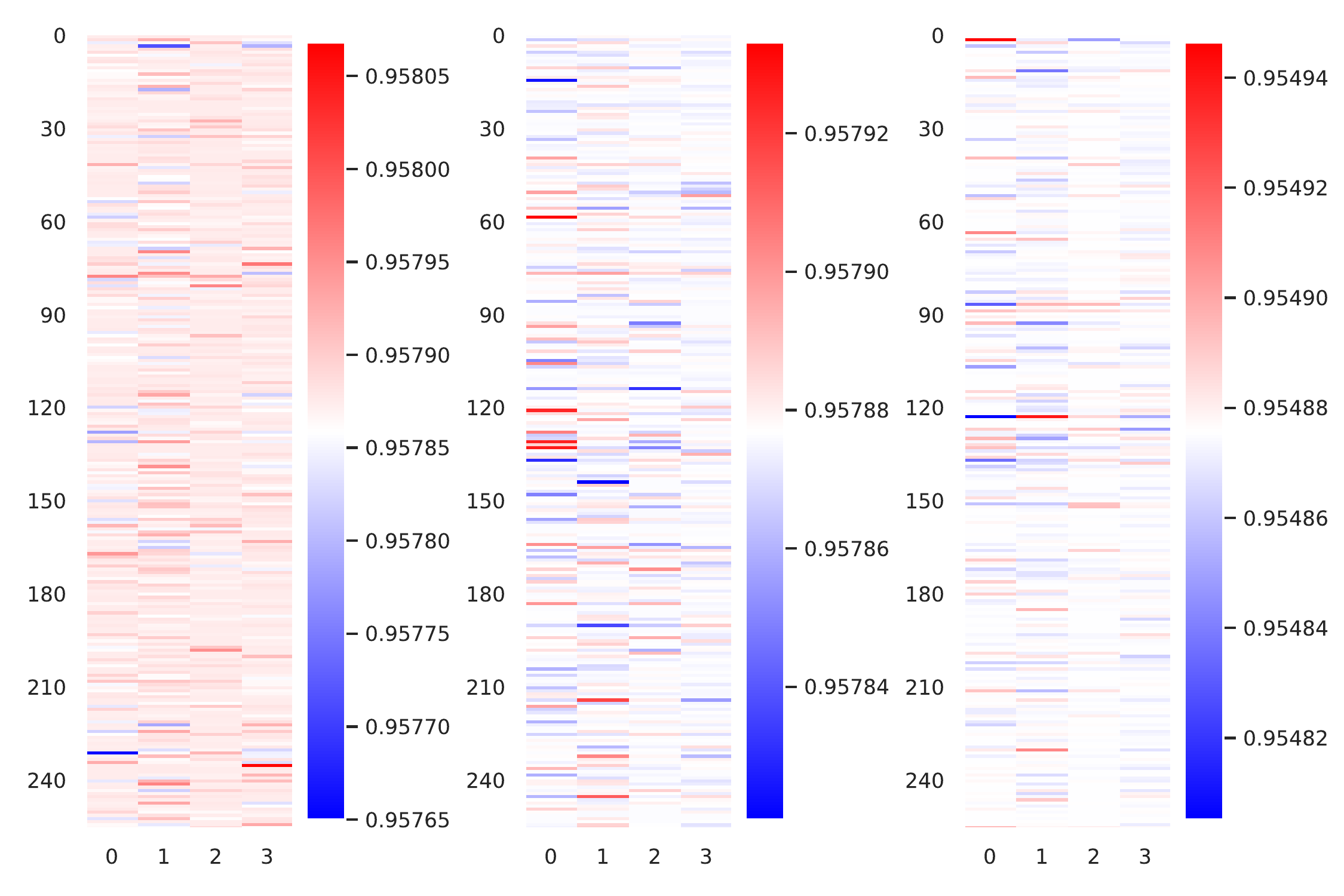}
    \end{minipage}
    }_{Stage1}$
   $\underbrace{    
    \begin{minipage}{0.25\linewidth}
        \includegraphics[width=1\linewidth]{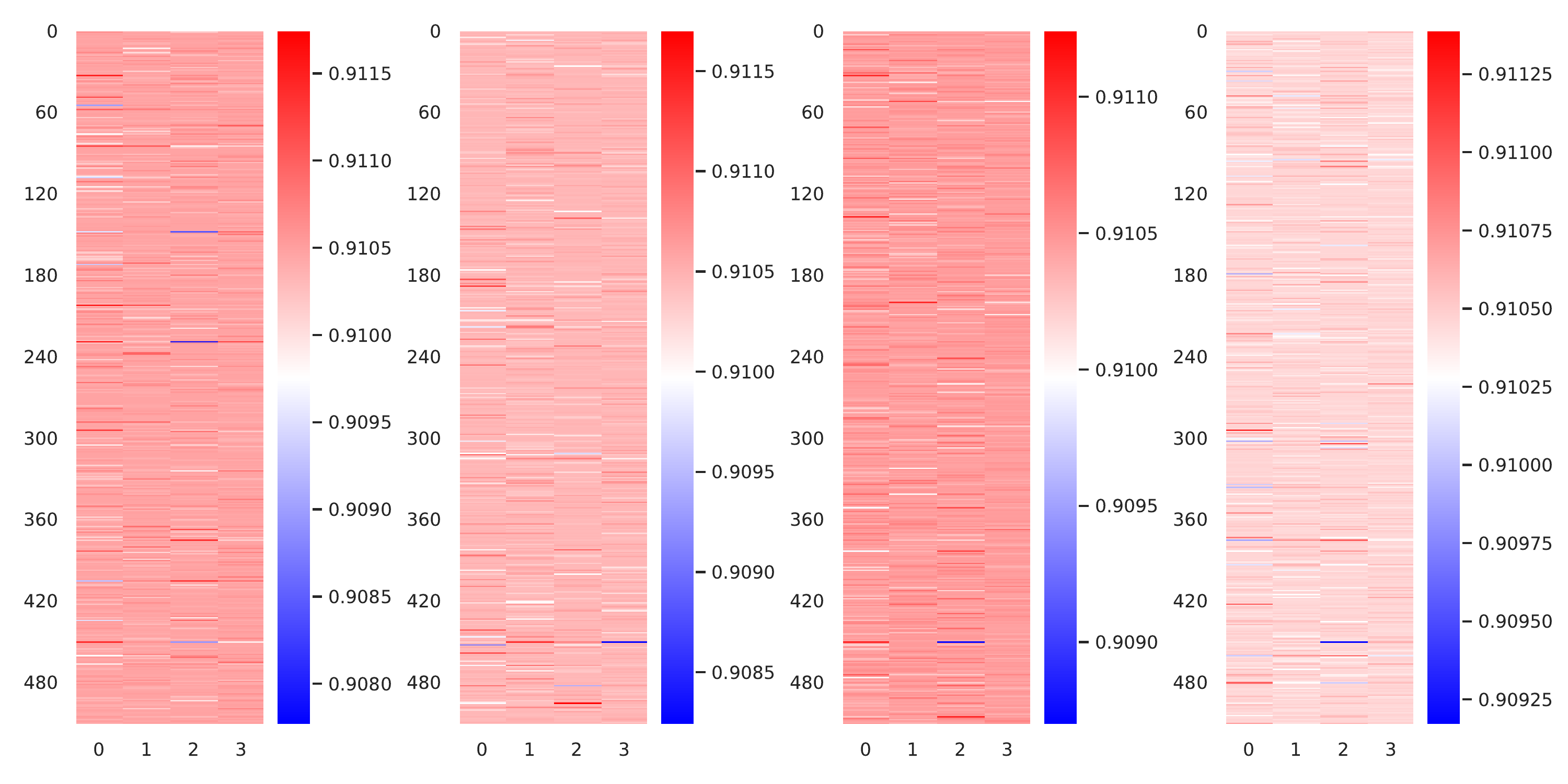}
    \end{minipage}
    }_{Stage2}$
   $\underbrace{
     \begin{minipage}{0.37\linewidth}
        \includegraphics[width=1\linewidth]{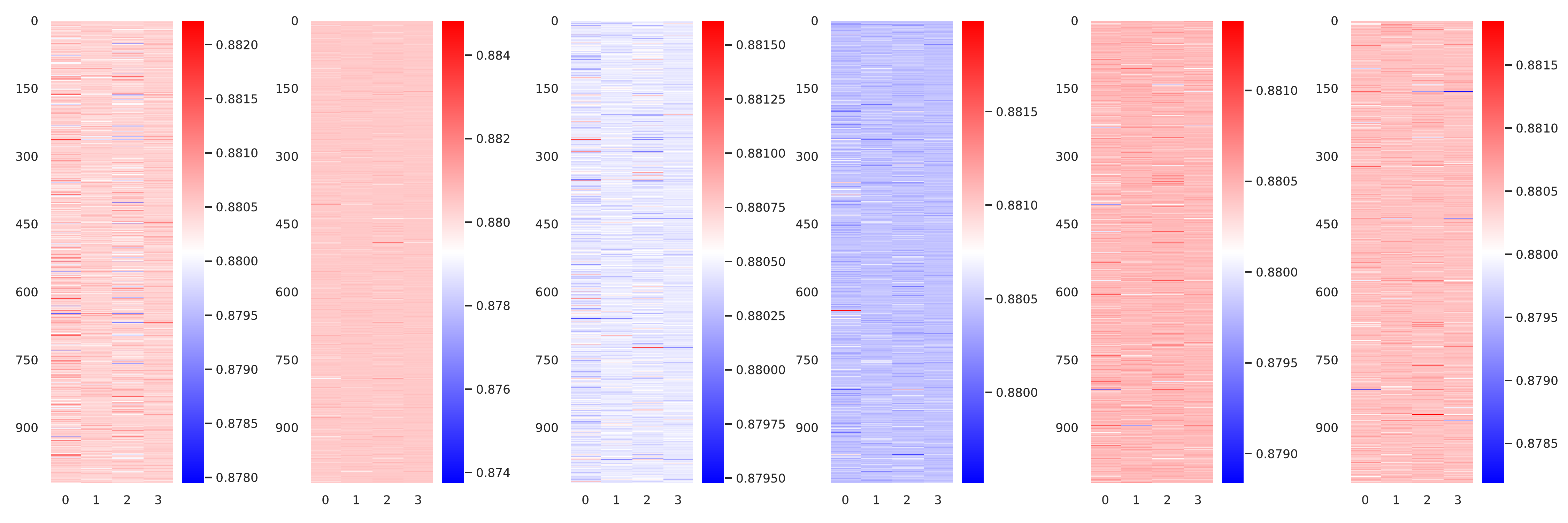}
    \end{minipage}
    }_{Stage3}$
   $\underbrace{
    \begin{minipage}{0.18\linewidth}
        \includegraphics[width=1\linewidth]{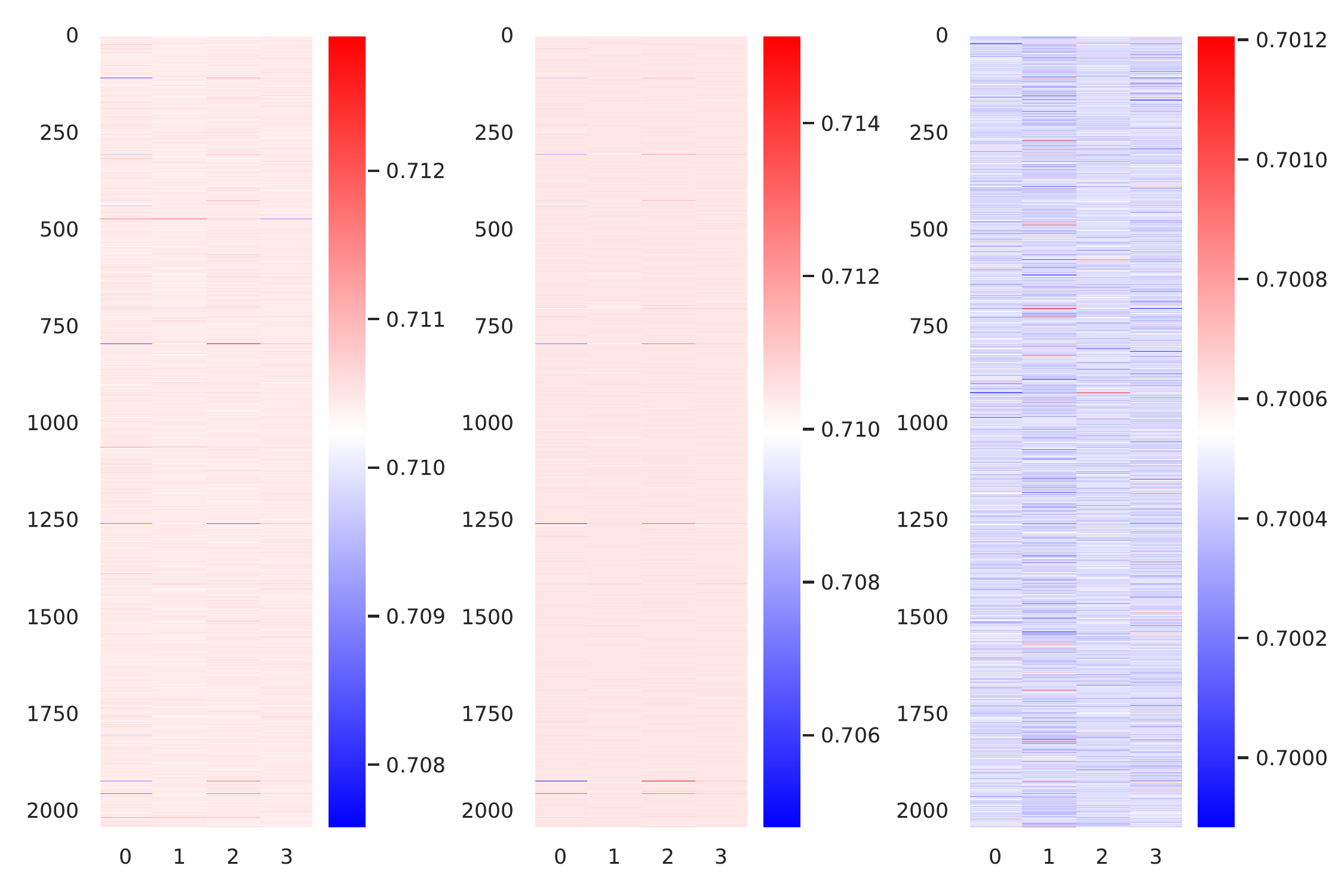}
    \end{minipage}
    }_{Stage4}$
}


\subfigure[P$\to$R (CDAN)]{
  \centering
  $\underbrace{
    \begin{minipage}{0.18\linewidth}
        \includegraphics[width=1\linewidth]{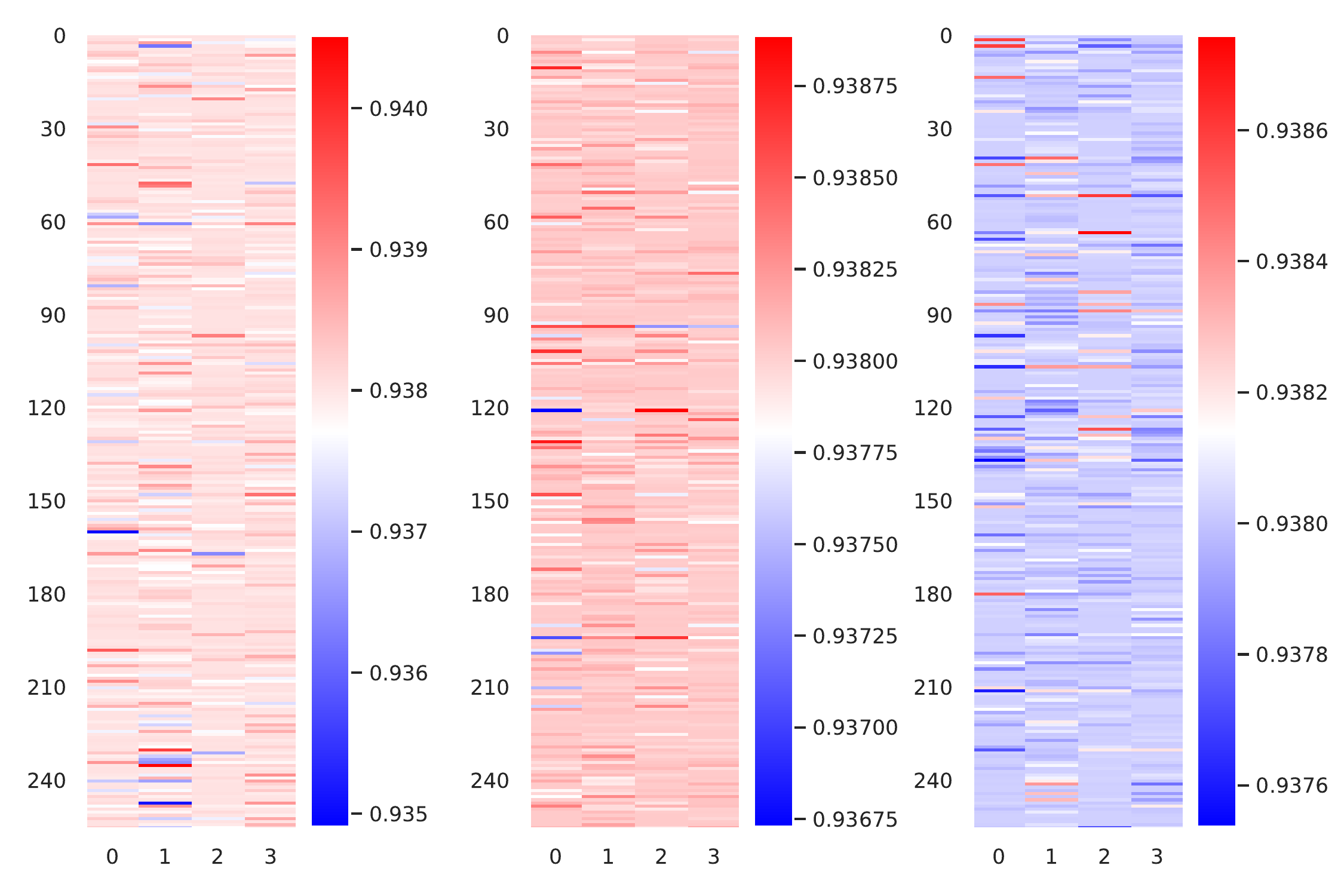}
    \end{minipage}}_{Stage1}$
  $\underbrace{    
    \begin{minipage}{0.25\linewidth}
        \includegraphics[width=1\linewidth]{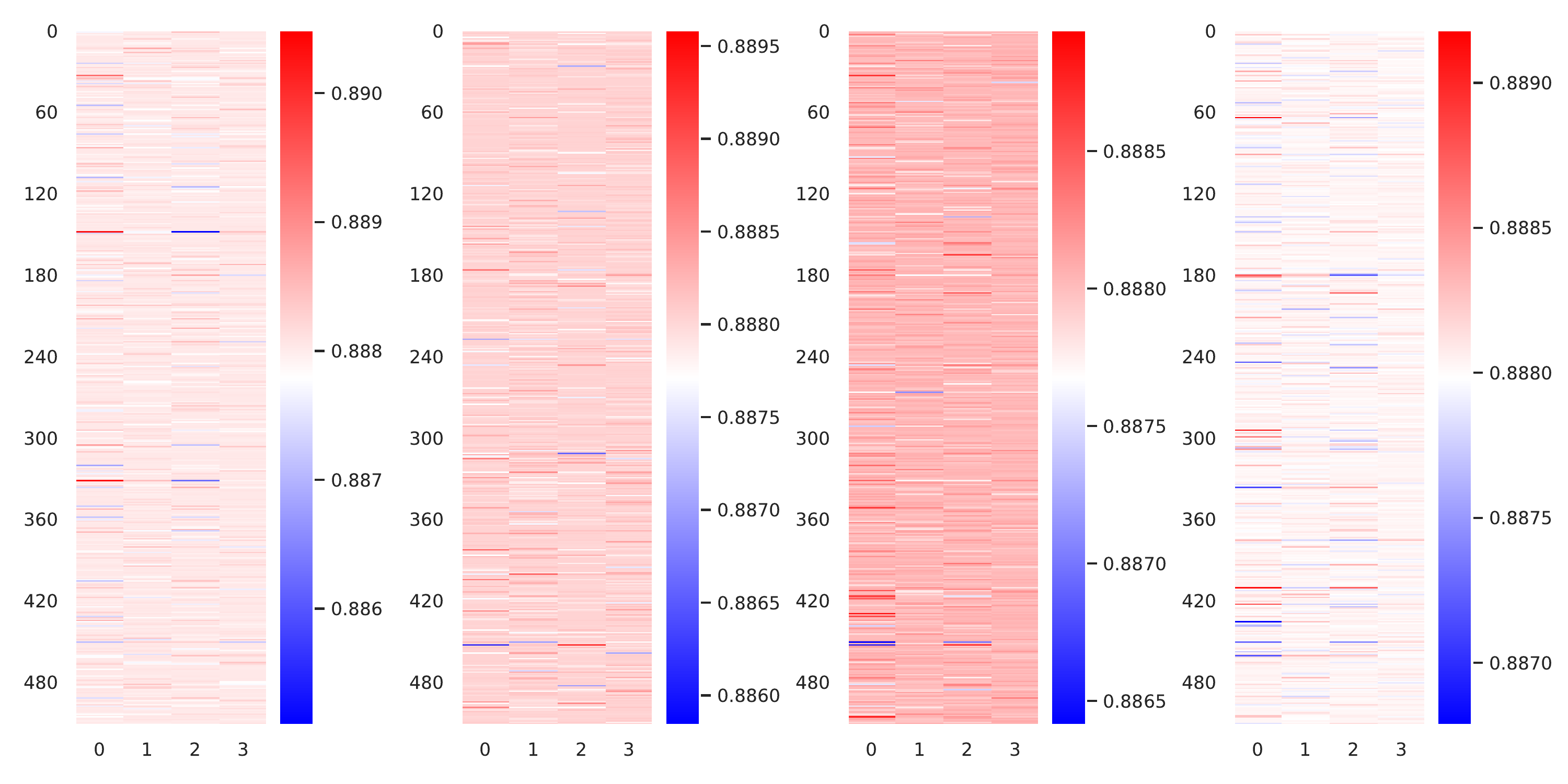}
    \end{minipage}}_{Stage2}$
    
  $\underbrace{
     \begin{minipage}{0.37\linewidth}
        \includegraphics[width=1\linewidth]{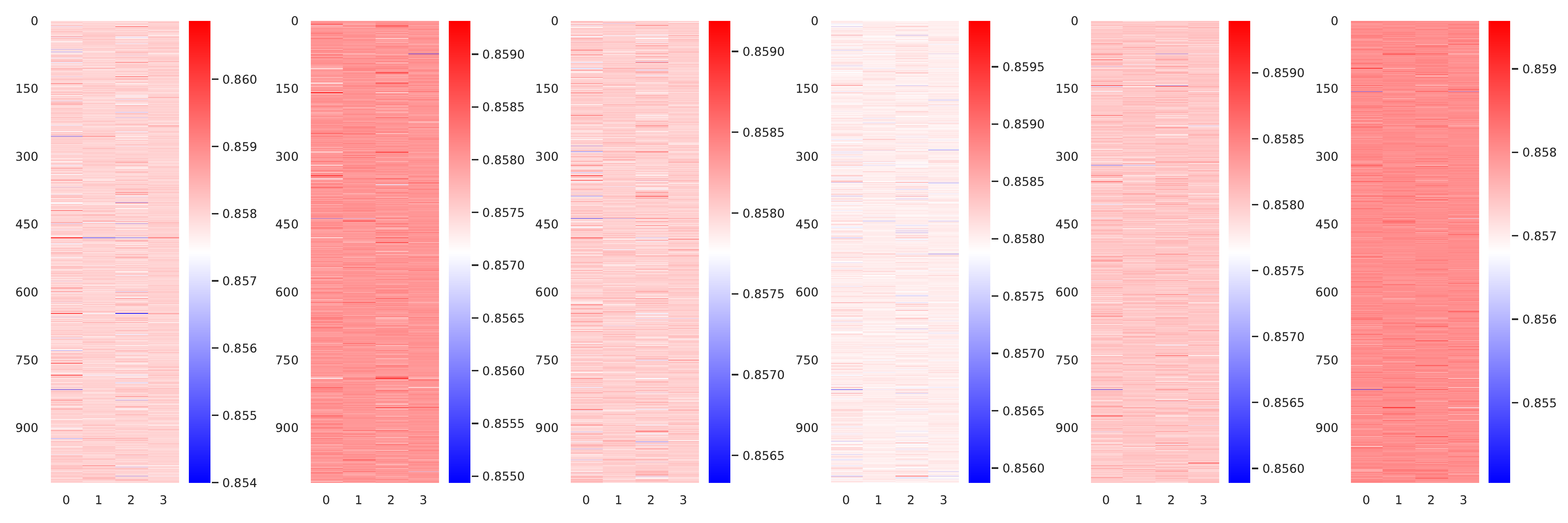}
    \end{minipage}}_{Stage3}$
  $\underbrace{
    \begin{minipage}{0.18\linewidth}
        \includegraphics[width=1\linewidth]{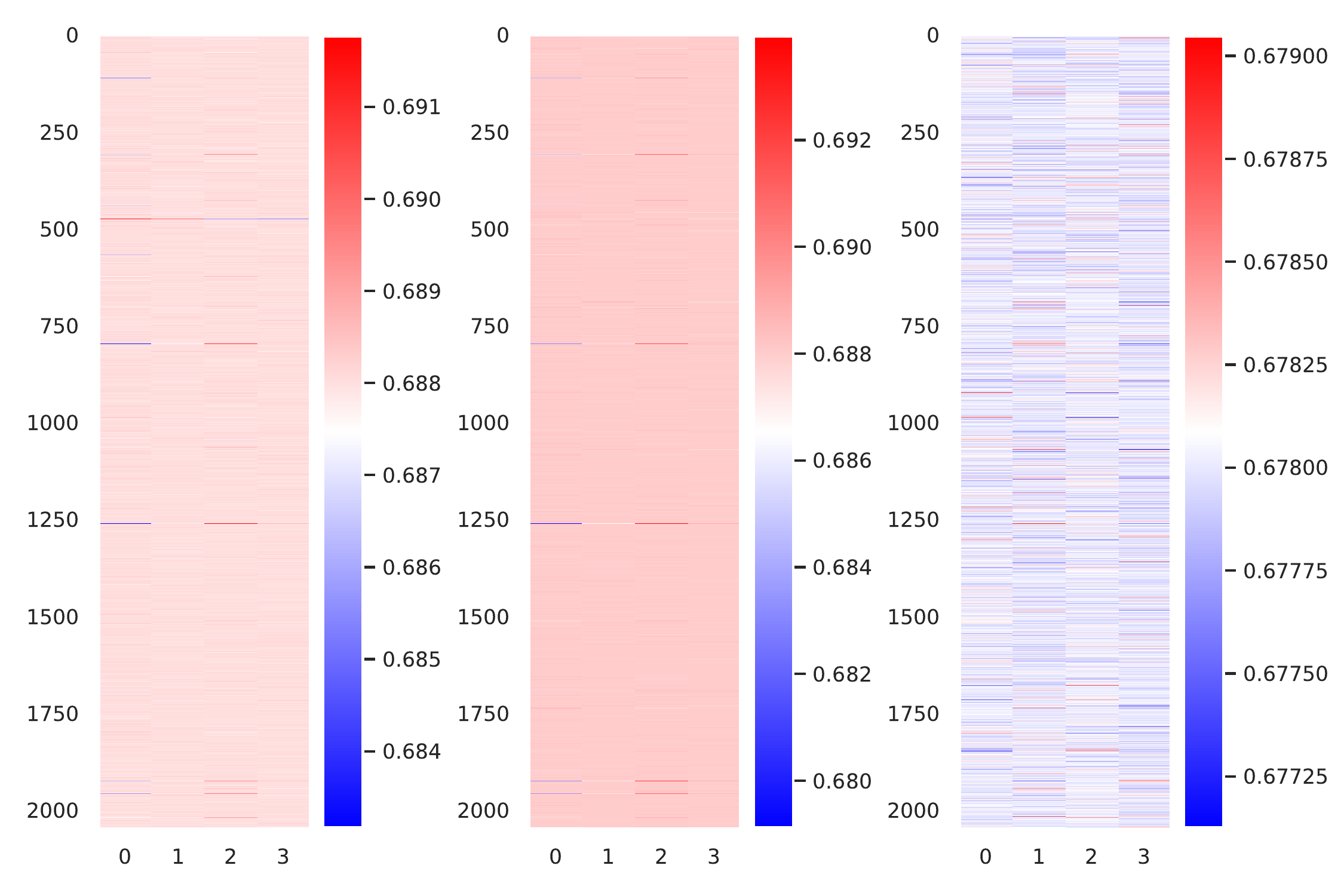}
    \end{minipage}}_{Stage4}$
}

    

\subfigure[P$\to$A (CDAN)]{
  \centering
  $\underbrace{
    \begin{minipage}{0.18\linewidth}
        \includegraphics[width=1\linewidth]{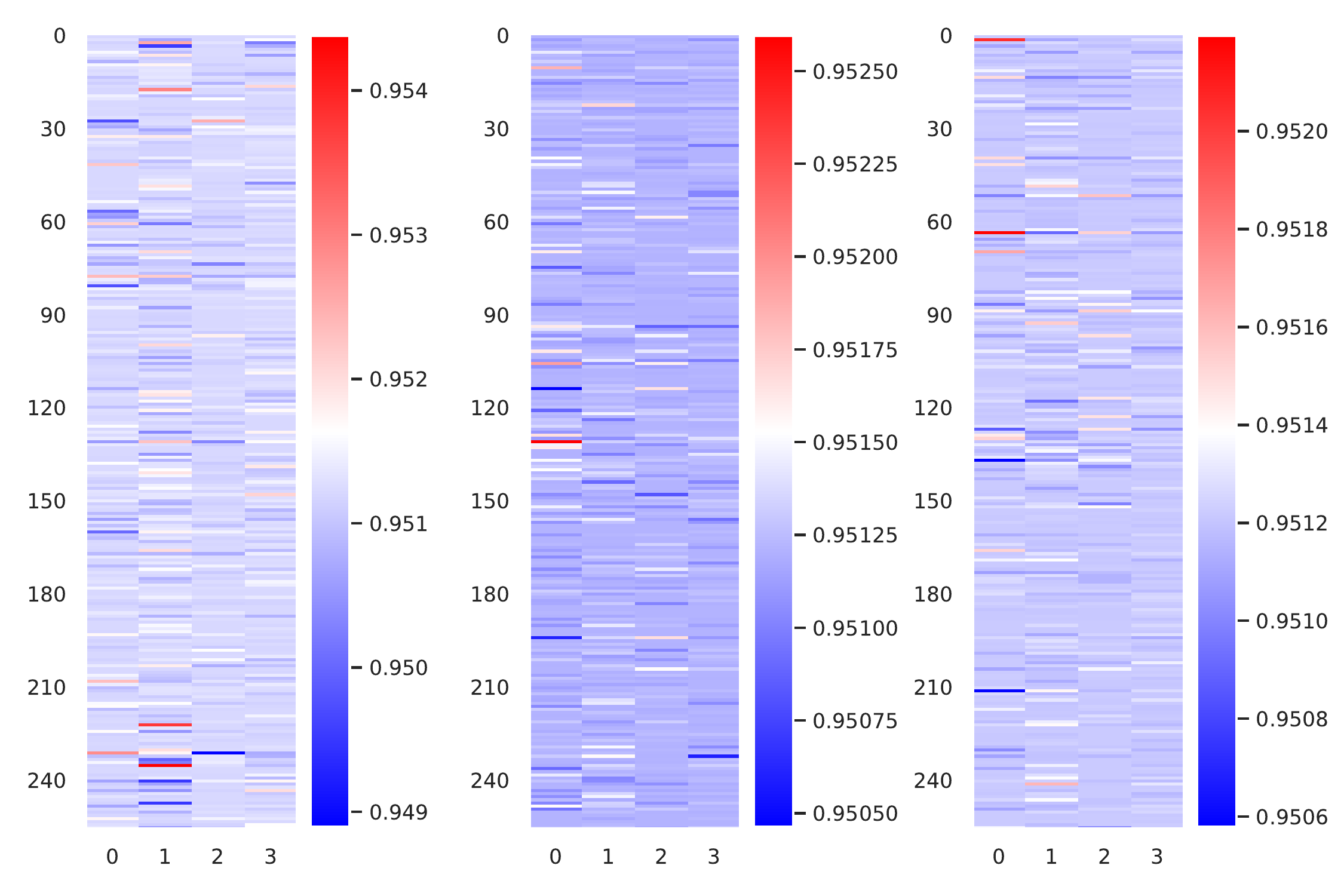}
    \end{minipage}}_{Stage1}$
  $\underbrace{    
    \begin{minipage}{0.25\linewidth}
        \includegraphics[width=1\linewidth]{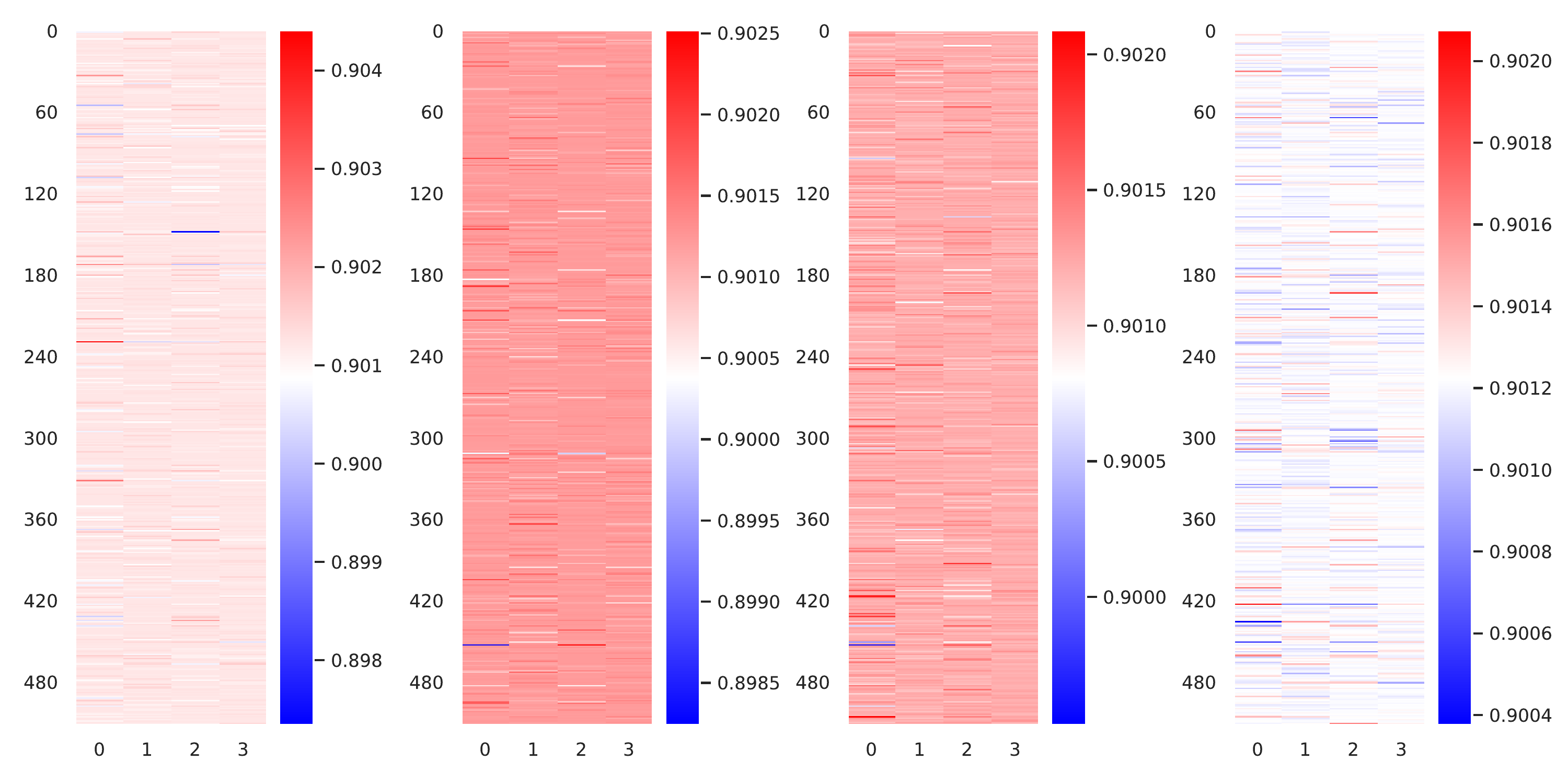}
    \end{minipage}}_{Stage2}$
    
  $\underbrace{
     \begin{minipage}{0.37\linewidth}
        \includegraphics[width=1\linewidth]{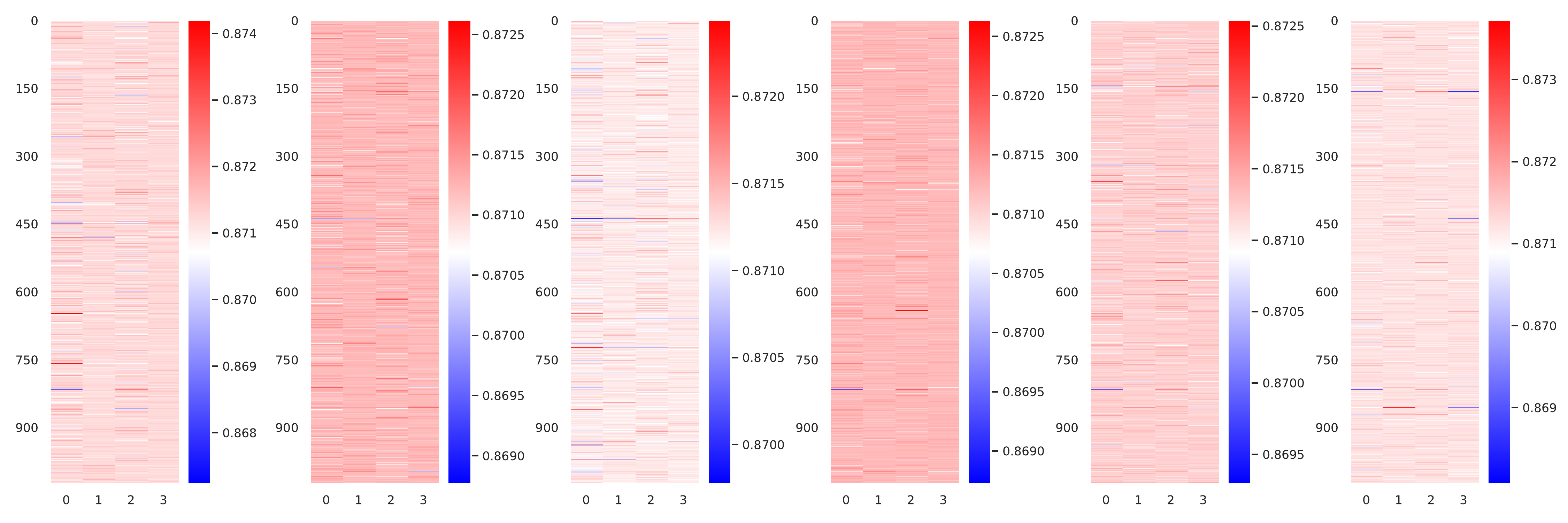}
    \end{minipage}}_{Stage3}$
  $\underbrace{
    \begin{minipage}{0.18\linewidth}
        \includegraphics[width=1\linewidth]{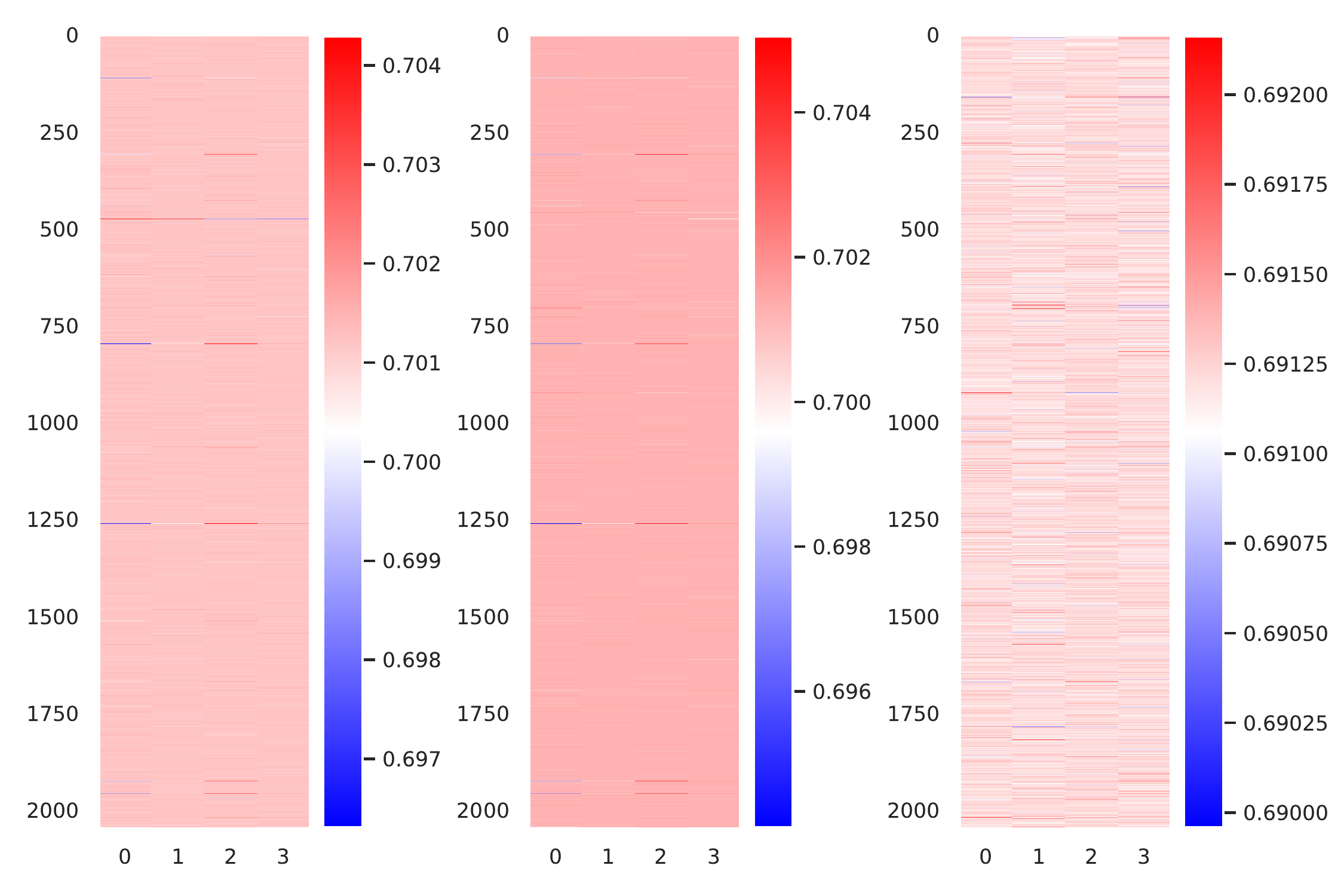}
    \end{minipage}}_{Stage4}$
}

    
    \caption{Heat-map visualization of RA-gates on three randomly selected adaptation tasks from Office-Home benchmarks. We use ResNet-50 as the backbone. These pictures are best viewed in the electronic version.}
    \label{subfig:Gate}
    \vspace{-2mm}
\end{figure*}
\begin{figure*}
  \label{analysis}
  \centering
  \subfigure[CDAN+BN]{
  \label{tsne:BN}
    \begin{minipage}{0.32\linewidth}
        \includegraphics[width=1\linewidth]{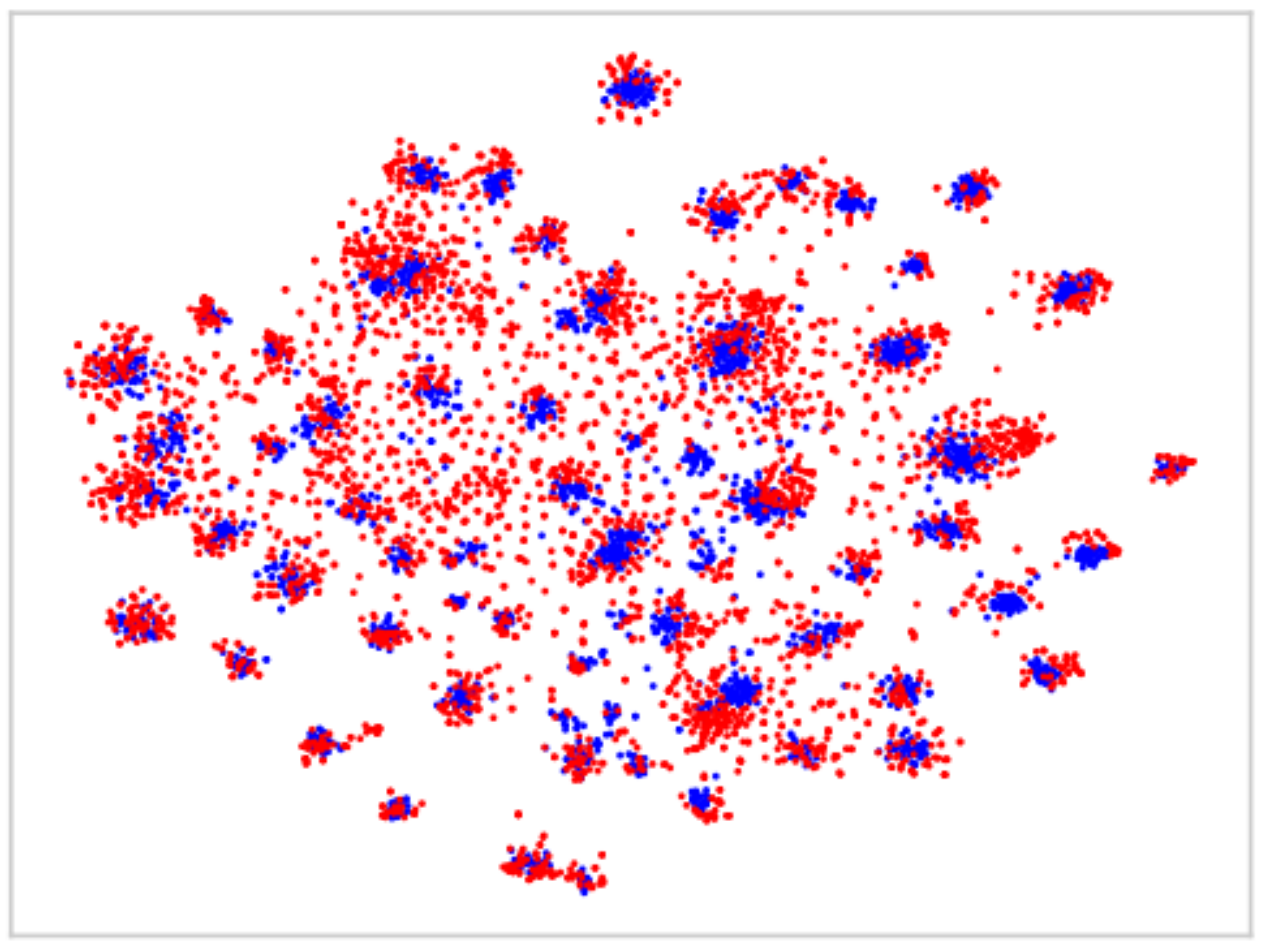}
    \end{minipage}}
 \subfigure[CDAN+AutoDIAL]{
  \label{tsne:AutoDIAL}
    \begin{minipage}{0.32\linewidth}
        \includegraphics[width=1\linewidth]{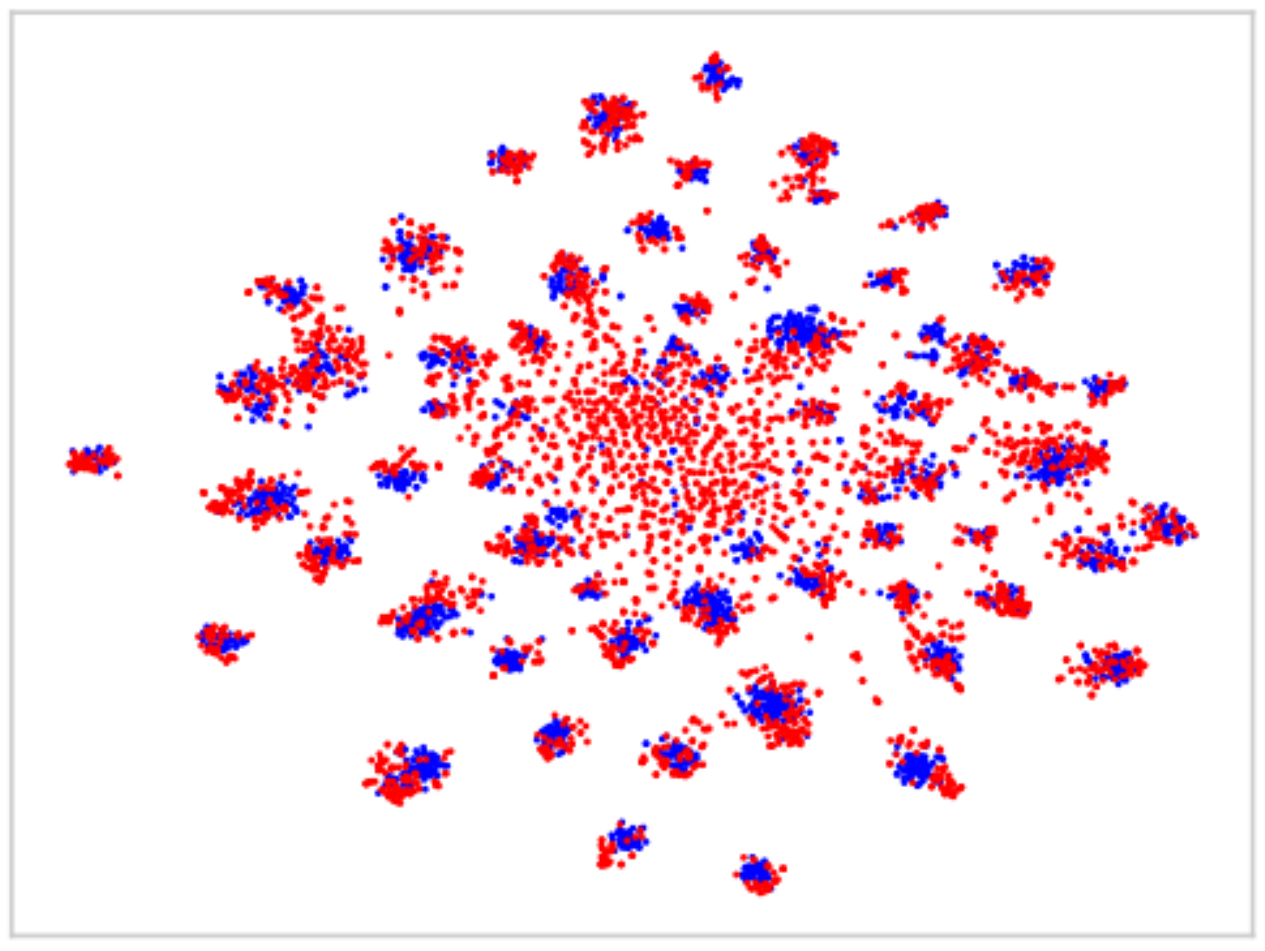}
    \end{minipage}}
  \subfigure[CDAN+DSBN]{
   \label{tsne:DSBN}
    \begin{minipage}{0.32\linewidth}
        \includegraphics[width=1\linewidth]{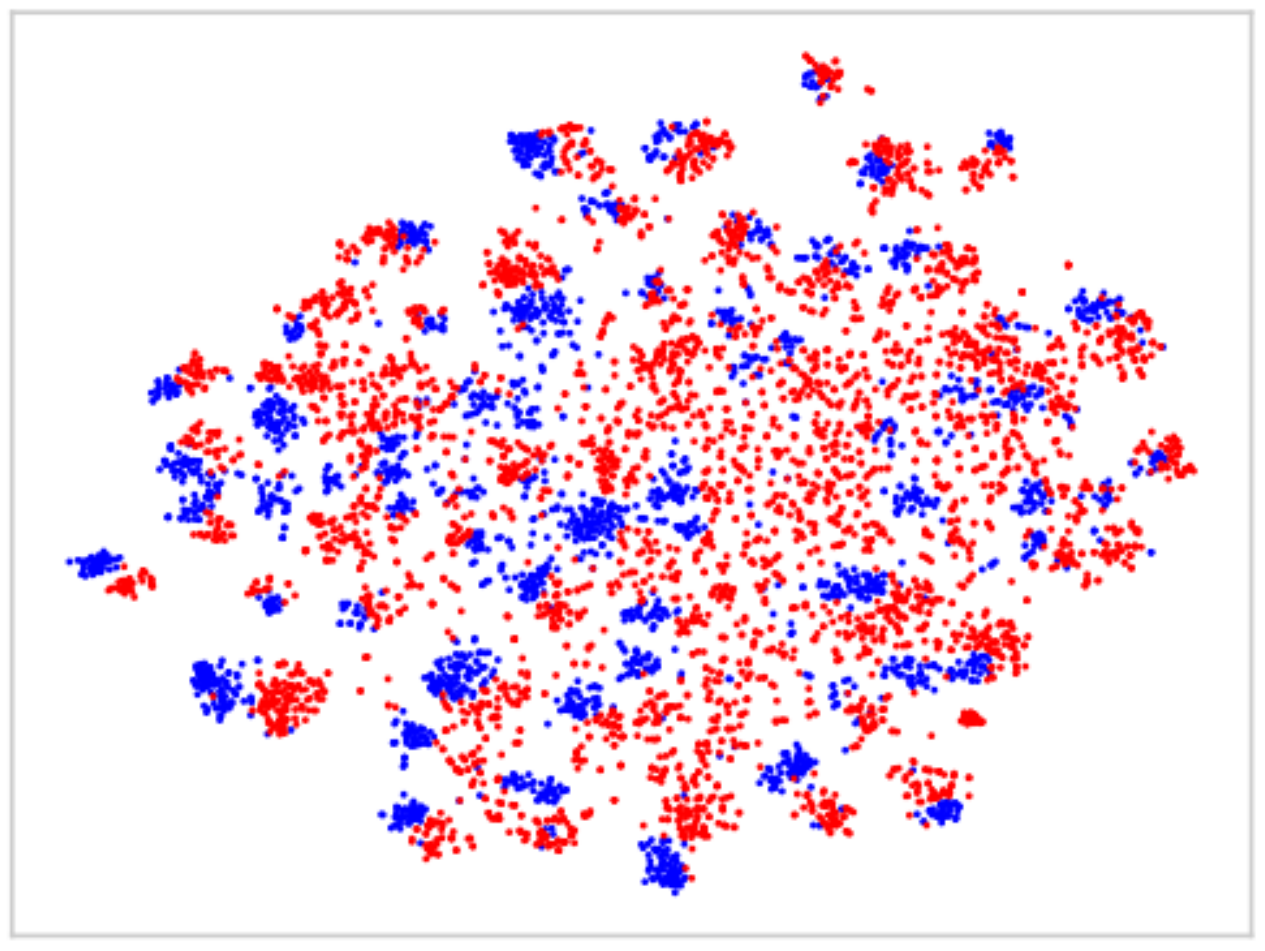}
    \end{minipage}}
    \subfigure[CDAN+TN]{
        \label{tsne:TN}
         \begin{minipage}{0.32\linewidth}
            \includegraphics[width=1\linewidth]{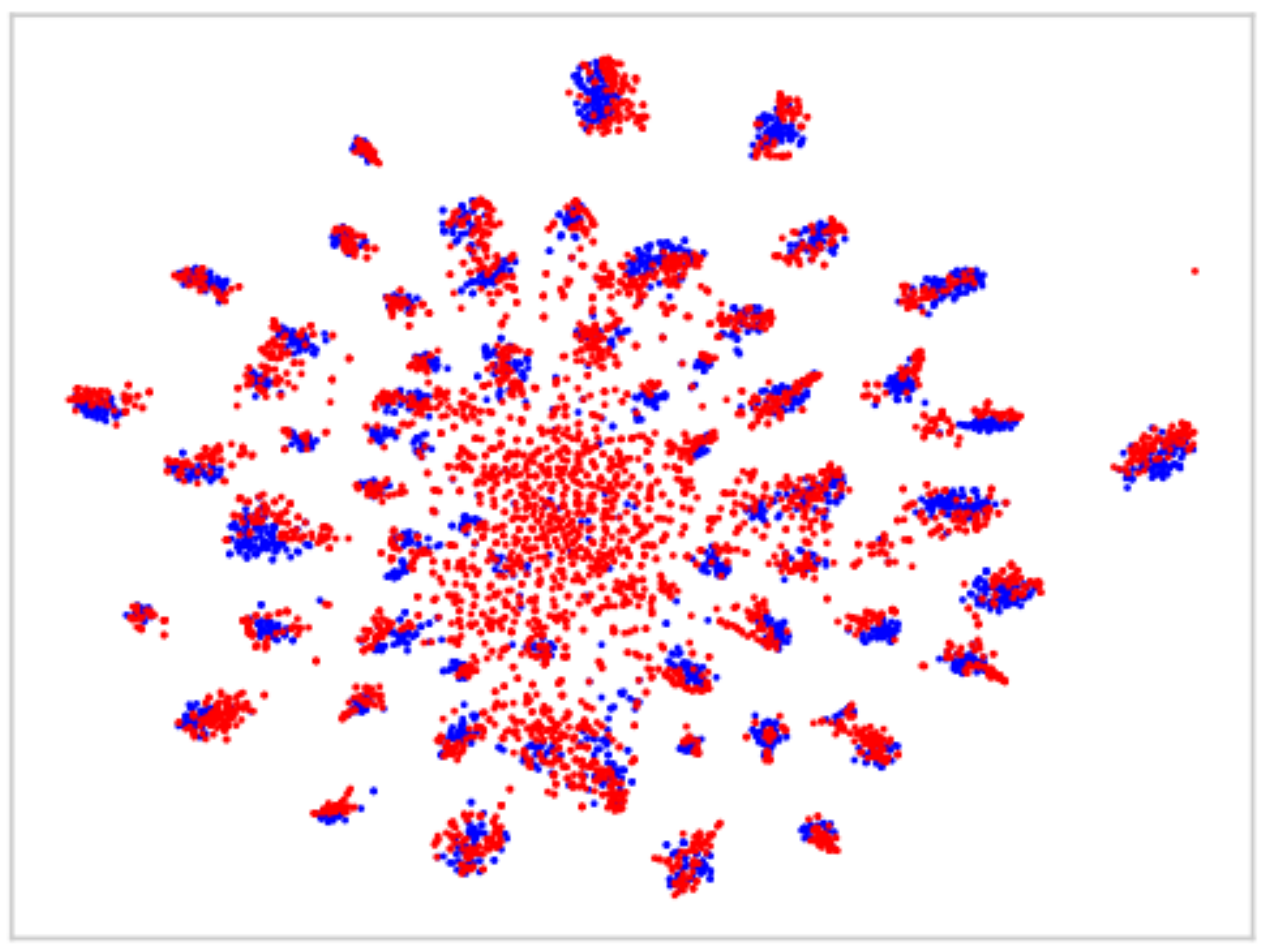}
     \end{minipage}
     }
      \subfigure[CDAN+RN]{
        \label{tsne:RN}
         \begin{minipage}{0.32\linewidth}
            \includegraphics[width=1\linewidth]{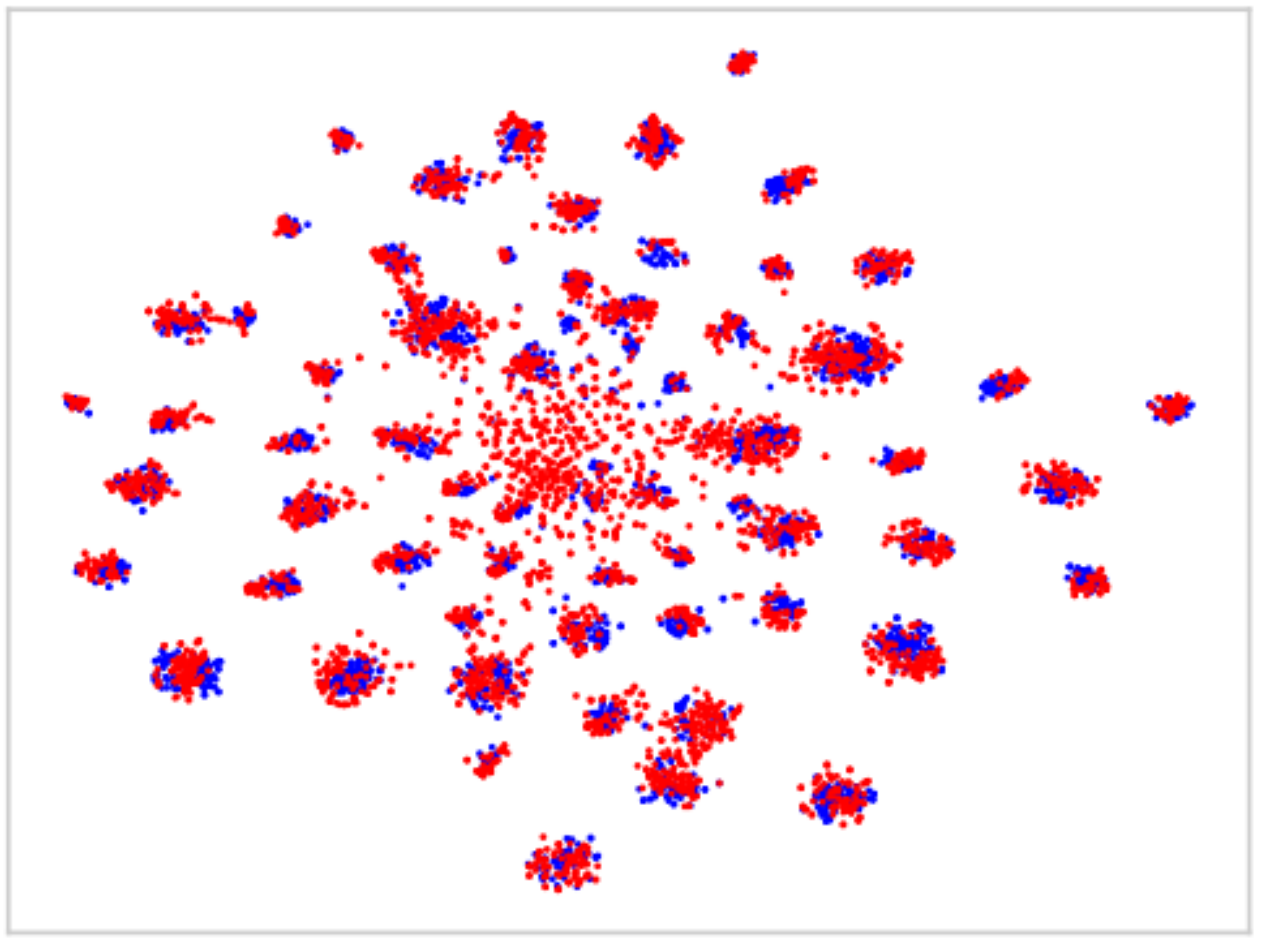}
     \end{minipage}
     }
    \label{TSNE}
    \caption{Visualization of features from the models with different normalization layers on the UDA task \textcolor{blue}{Art (Source)} $\to$ \textcolor{red}{Clipart (Target)} (Office-Home).}
    \vspace{-2mm}
\end{figure*}

\subsection{Additional Visualization}
\label{sup_sec:Visualization}

\subsubsection{Visualization of $g$ in RA}
For better understanding of RA, we also show the visualization of $g$ in RA on UDA tasks Art $\to$ Clipart of Office-Home. We show the last RN of each layer (\ie, \textit{bn3} in each layer of ResNet50) due to the page limitation. We refer to each stage as stage $1$, $2$, $3$ and $4$ with the channel numbers as $256$, $512$, $1024$, and $2048$.
The ``1, 2, 3, 4'' on the abscissa axis denote the source mean, source variance, target mean, and target variance, respectively.
The ordinates denote the values of $g$.

As illustrated in Fig.~\ref{subfig:Gate}, RA 
conducts the domain alignment at the intermediate layers in different ways automatically.
Note that, as the number of channels increases, the weights of gates become smaller, which is consistent with the conclusion that the different transferability in the various layers in ~\cite{yosinski2014transferable}, and ensures the significance of RA. 
Similar observations can also be found in other DA scenarios.


\subsubsection{Feature Visualization}
To further understand the effectiveness of the proposed RN, we visualize the learned representation spaces of different feature normalization modules: vanilla BN~\cite{ioffe2015BN}, AutoDIAL~\cite{cariucci2017autodial}, DSBN~\cite{chang2019domain}, TN~\cite{wang2019TN}, and the proposed RN.
We leverage t-SNE~\cite{tsne} to visualize the feature representations in the bottleneck layer of ResNet-50.

As shown in Fig.~\ref{TSNE}, we notice that CDAN+DSBN suffers from negative transfer, which is the main reason for the sub-optimal results (see Table~\ref{tab:ImageCLEFDA_results}, Table~\ref{tab:OfficeHome_results}, Table~\ref{tab:ResultsOnVisDA2017}, and Table~\ref{tab:OfficeHome_Results_MSDA}).
We also observe that the source and target representations are aligned better by the models integrated RN, compared with existing normalization counterparts.
It demonstrates that our RN is effective to learn the domain-invariant information.
Meanwhile, the cluster centers of two domains in the same class are closer, indicating that the greater ability of RN to learn the discriminative features.


\section{Conclusion}
In this paper, we propose a novel normalization layer for domain adaptation, termed Reciprocal Normalization (RN). We devise RN to address the problem that losing the domain information due to the misalignment of channels across domains. The proposed RN structurally aligns the source and target domains by conducting reciprocity across domains. As a generic alternative to BN, our RN can be easily applied to mainstream domain adaptation approaches. Extensive experiments on three benchmarks and three typical adaptation tasks validate that: i) the proposed RN outperforms existing normalization techniques in the context of domain adaptation; ii) popular domain adaptation approaches consistently benefit from our RN and obtain better classification performance on the target domain. 

For future work, we will attempt to reveal the theoretical insights within our RN and verify its versatility in other domain adaptation tasks (\eg, object detection~\cite{xu2020cross}, text detection~\cite{hou2020ham}, semantic segmentation~\cite{lv2020cross}, and person re-identification~\cite{liu2019adaptive}).

\bibliographystyle{ieee_fullname}
\bibliography{MAIN_for_review_TIP}

\begin{thebibliography}{10}\itemsep=-1pt

\bibitem{ba2016LN}
Jimmy~Lei Ba, Jamie~Ryan Kiros, and Geoffrey~E Hinton.
\newblock Layer normalization.
\newblock In {\em Advances in Neural Information Processing Systems}, 2016.

\bibitem{ben2010theory}
Shai Ben-David, John Blitzer, Koby Crammer, Alex Kulesza, Fernando Pereira, and
  Jennifer~Wortman Vaughan.
\newblock A theory of learning from different domains.
\newblock {\em Machine learning}, 2010.

\bibitem{bronskill2020tasknorm}
John Bronskill, Jonathan Gordon, James Requeima, Sebastian Nowozin, and Richard
  Turner.
\newblock Tasknorm: Rethinking batch normalization for meta-learning.
\newblock In {\em International Conference on Machine Learning}, pages
  1153--1164, 2020.

\bibitem{cao2018SAN}
Zhangjie Cao, Mingsheng Long, Jianmin Wang, and Michael~I Jordan.
\newblock Partial transfer learning with selective adversarial networks.
\newblock In {\em Proceedings of the IEEE/CVF Conference on Computer Vision and
  Pattern Recognition}, 2018.

\bibitem{cao2019ETN}
Zhangjie Cao, Kaichao You, Mingsheng Long, Jianmin Wang, and Qiang Yang.
\newblock Learning to transfer examples for partial domain adaptation.
\newblock In {\em Proceedings of the IEEE/CVF Conference on Computer Vision and
  Pattern Recognition}, 2019.

\bibitem{cariucci2017autodial}
F.~M. {Cariucci}, L. {Porzi}, B. {Caputo}, E. {Ricci}, and S.~R. {Bulò}.
\newblock Autodial: Automatic domain alignment layers.
\newblock In {\em Proceedings of the IEEE International Conference on Computer
  Vision}, 2017.

\bibitem{chang2019domain}
Woong-Gi Chang, Tackgeun You, Seonguk Seo, Suha Kwak, and Bohyung Han.
\newblock Domain-specific batch normalization for unsupervised domain
  adaptation.
\newblock In {\em Proceedings of the IEEE/CVF Conference on Computer Vision and
  Pattern Recognition}, 2019.

\bibitem{Chen_2019_CVPR}
Chaoqi Chen, Weiping Xie, Wenbing Huang, Yu Rong, Xinghao Ding, Yue Huang,
  Tingyang Xu, and Junzhou Huang.
\newblock Progressive feature alignment for unsupervised domain adaptation.
\newblock In {\em Proceedings of the IEEE/CVF Conference on Computer Vision and
  Pattern Recognition}, 2019.

\bibitem{chen2020structure}
Qingchao Chen and Yang Liu.
\newblock Structure-aware feature fusion for unsupervised domain adaptation.
\newblock In {\em Proceedings of the AAAI Conference on Artificial
  Intelligence}, pages 10567--10574, 2020.

\bibitem{chen2019BSP}
Xinyang Chen, Sinan Wang, Mingsheng Long, and Jianmin Wang.
\newblock Transferability vs. discriminability: Batch spectral penalization for
  adversarial domain adaptation.
\newblock In {\em International Conference on Machine Learning}, 2019.

\bibitem{cui2020BNM}
Shuhao Cui, Shuhui Wang, Junbao Zhuo, Liang Li, Qingming Huang, and Qi Tian.
\newblock Towards discriminability and diversity: Batch nuclear-norm
  maximization under label insufficient situations.
\newblock In {\em Proceedings of the IEEE/CVF Conference on Computer Vision and
  Pattern Recognition}, 2020.

\bibitem{cui2020gvb}
Shuhao Cui, Shuhui Wang, Junbao Zhuo, Chi Su, Qingming Huang, and Tian Qi.
\newblock Gradually vanishing bridge for adversarial domain adaptation.
\newblock In {\em Proceedings of the IEEE/CVF Conference on Computer Vision and
  Pattern Recognition}, 2020.

\bibitem{du2020metanorm}
Yingjun Du, Xiantong Zhen, Ling Shao, and Cees~GM Snoek.
\newblock Metanorm: Learning to normalize few-shot batches across domains.
\newblock In {\em International Conference on Learning Representations}, 2020.

\bibitem{fatras2021JUMBOT}
Kilian Fatras, Thibault S{\'e}journ{\'e}, R{\'e}mi Flamary, and Nicolas Courty.
\newblock Unbalanced minibatch optimal transport; applications to domain
  adaptation.
\newblock In {\em International Conference on Machine Learning}, pages
  3186--3197, 2021.

\bibitem{ganin2015DANN}
Yaroslav Ganin and Victor~S. Lempitsky.
\newblock Unsupervised domain adaptation by backpropagation.
\newblock In {\em International Conference on Machine Learning}, 2015.

\bibitem{gao2021representative}
Shang-Hua Gao, Qi Han, Duo Li, Ming-Ming Cheng, and Pai Peng.
\newblock Representative batch normalization with feature calibration.
\newblock In {\em Proceedings of the IEEE/CVF Conference on Computer Vision and
  Pattern Recognition}, pages 8669--8679, 2021.

\bibitem{greron2007mmd}
Arthur Gretton, Karsten Borgwardt, Malte Rasch, Bernhard Sch\"{o}lkopf, and
  Alex~J. Smola.
\newblock A kernel method for the two-sample-problem.
\newblock In {\em Advances in Neural Information Processing Systems}, 2007.

\bibitem{he2016resnet}
Kaiming He, Xiangyu Zhang, Shaoqing Ren, and Jian Sun.
\newblock Deep residual learning for image recognition.
\newblock In {\em Proceedings of the IEEE/CVF Conference on Computer Vision and
  Pattern Recognition}, 2016.

\bibitem{hou2020ham}
Jie-Bo Hou, Xiaobin Zhu, Chang Liu, Kekai Sheng, Long-Huang Wu, Hongfa Wang,
  and Xu-Cheng Yin.
\newblock Ham: Hidden anchor mechanism for scene text detection.
\newblock {\em IEEE Transactions on Image Processing}, pages 7904--7916, 2020.

\bibitem{ioffe2015BN}
Sergey Ioffe and Christian Szegedy.
\newblock Batch normalization: Accelerating deep training by reducing internal
  covariate shift.
\newblock In {\em International Conference on Machine Learning}, 2015.

\bibitem{jiang2020implicit}
Xiang Jiang, Qicheng Lao, Stan Matwin, and Mohammad Havaei.
\newblock Implicit class-conditioned domain alignment for unsupervised domain
  adaptation.
\newblock In {\em International Conference on Machine Learning}, 2020.

\bibitem{kang2019CAN}
Guoliang Kang, Lu Jiang, Yi Yang, and Alexander~G Hauptmann.
\newblock Contrastive adaptation network for unsupervised domain adaptation.
\newblock In {\em Proceedings of the IEEE/CVF Conference on Computer Vision and
  Pattern Recognition}, 2019.

\bibitem{kurmi2019CADA}
Vinod~Kumar Kurmi, Shanu Kumar, and Vinay~P Namboodiri.
\newblock Attending to discriminative certainty for domain adaptation.
\newblock In {\em Proceedings of the IEEE/CVF Conference on Computer Vision and
  Pattern Recognition}, pages 491--500, 2019.

\bibitem{lee2019sliced}
Chen-Yu Lee, Tanmay Batra, Mohammad~Haris Baig, and Daniel Ulbricht.
\newblock Sliced wasserstein discrepancy for unsupervised domain adaptation.
\newblock In {\em Proceedings of the IEEE/CVF Conference on Computer Vision and
  Pattern Recognition}, 2019.

\bibitem{Lee2019DTA}
Seungmin Lee, Dongwan Kim, Namil Kim, and Seong-Gyun Jeong.
\newblock Drop to adapt: Learning discriminative features for unsupervised
  domain adaptation.
\newblock In {\em Proceedings of the IEEE International Conference on Computer
  Vision}, 2019.

\bibitem{li2020online}
Da Li and Timothy Hospedales.
\newblock Online meta-learning for multi-source and semi-supervised domain
  adaptation.
\newblock In {\em European Conference on Computer Vision}, pages 382--403,
  2020.

\bibitem{li2020DCAN}
Shuang Li, Chi~Harold Liu, Qiuxia Lin, Binhui Xie, Zhengming Ding, Gao Huang,
  and Jian Tang.
\newblock Domain conditioned adaptation network.
\newblock In {\em Proceedings of the AAAI Conference on Artificial
  Intelligence}, 2020.

\bibitem{Li21BCDM}
Shuang Li, Fangrui Lv, Binhui Xie, Chi~Harold Liu, Jian Liang, and Chen Qin.
\newblock Bi-classifier determinacy maximization for unsupervised domain
  adaptation.
\newblock In {\em Proceedings of the AAAI Conference on Artificial
  Intelligence}, 2021.

\bibitem{li2019ConvNorm}
Yunsheng Li and Nuno Vasconcelos.
\newblock Efficient multi-domain learning by covariance normalization.
\newblock In {\em Proceedings of the IEEE/CVF Conference on Computer Vision and
  Pattern Recognition}, pages 5424--5433, 2019.

\bibitem{li2017AdaBN}
Yanghao Li, Naiyan Wang, Jianping Shi, Jiaying Liu, and Xiaodi Hou.
\newblock Revisiting batch normalization for practical domain adaptation.
\newblock In {\em International Conference on Learning Representations}, 2017.

\bibitem{liang2020shot}
Jian Liang, Dapeng Hu, and Jiashi Feng.
\newblock Do we really need to access the source data? source hypothesis
  transfer for unsupervised domain adaptation.
\newblock In {\em International Conference on Machine Learning}, 2020.

\bibitem{liang2021shot}
Jian Liang, Dapeng Hu, Yunbo Wang, Ran He, and Jiashi Feng.
\newblock Source data-absent unsupervised domain adaptation through hypothesis
  transfer and labeling transfer.
\newblock {\em IEEE Transactions on Pattern Analysis and Machine Intelligence},
  2021.

\bibitem{liang2020BA3US}
Jian Liang, Yunbo Wang, Dapeng Hu, Ran He, and Jiashi Feng.
\newblock A balanced and uncertainty-aware approach for partial domain
  adaptation.
\newblock In {\em Proceedings of the European Conference on Computer Vision},
  2020.

\bibitem{liu2020EvoNorm}
Hanxiao Liu, Andy Brock, Karen Simonyan, and Quoc Le.
\newblock Evolving normalization-activation layers.
\newblock In {\em Advances in Neural Information Processing Systems}, pages
  13539--13550, 2020.

\bibitem{liu2019adaptive}
Jiawei Liu, Zheng-Jun Zha, Di Chen, Richang Hong, and Meng Wang.
\newblock Adaptive transfer network for cross-domain person re-identification.
\newblock In {\em Proceedings of the IEEE/CVF Conference on Computer Vision and
  Pattern Recognition}, 2019.

\bibitem{long2015DAN}
Mingsheng Long, Yue Cao, Jianmin Wang, and Michael Jordan.
\newblock Learning transferable features with deep adaptation networks.
\newblock In {\em International Conference on Machine Learning}, 2015.

\bibitem{long2018CDAN}
Mingsheng Long, Zhangjie Cao, Jianmin Wang, and Michael~I Jordan.
\newblock Conditional adversarial domain adaptation.
\newblock In {\em Advances in Neural Information Processing Systems}, 2018.

\bibitem{long2017JAN}
Mingsheng Long, Han Zhu, Jianmin Wang, and Michael~I Jordan.
\newblock Deep transfer learning with joint adaptation networks.
\newblock In {\em International Conference on Machine Learning}, 2017.

\bibitem{luo2018SN}
Ping Luo, Jiamin Ren, Zhanglin Peng, Ruimao Zhang, and Jingyu Li.
\newblock Differentiable learning-to-normalize via switchable normalization.
\newblock In {\em International Conference on Learning Representations}, 2018.

\bibitem{lv2020cross}
Fengmao Lv, Tao Liang, Xiang Chen, and Guosheng Lin.
\newblock Cross-domain semantic segmentation via domain-invariant interactive
  relation transfer.
\newblock In {\em Proceedings of the IEEE/CVF Conference on Computer Vision and
  Pattern Recognition}, pages 4334--4343, 2020.

\bibitem{mancini2018DAlayers}
Massimilano Mancini, Lorenzo Porzi, Samuel Rota~Bul\`o, Barbara Caputo, and
  Elisa Ricci.
\newblock Boosting domain adaptation by discovering latent domains.
\newblock In {\em Proceedings of the IEEE/CVF Conference on Computer Vision and
  Pattern Recognition}, 2018.

\bibitem{Murez_2018_CVPR}
Zak Murez, Soheil Kolouri, David Kriegman, Ravi Ramamoorthi, and Kyungnam Kim.
\newblock Image to image translation for domain adaptation.
\newblock In {\em Proceedings of the IEEE/CVF Conference on Computer Vision and
  Pattern Recognition}, 2018.

\bibitem{paszke2019pytorch}
Adam Paszke, Sam Gross, Francisco Massa, Adam Lerer, James Bradbury, Gregory
  Chanan, Trevor Killeen, Zeming Lin, Natalia Gimelshein, Luca Antiga, et~al.
\newblock Pytorch: An imperative style, high-performance deep learning library.
\newblock In {\em Advances in Neural Information Processing Systems}, 2019.

\bibitem{peng2017visda}
Xingchao Peng, Ben Usman, Neela Kaushik, Judy Hoffman, Dequan Wang, and Kate
  Saenko.
\newblock Visda: The visual domain adaptation challenge.
\newblock {\em arXiv preprint arXiv:1710.06924}, 2017.

\bibitem{Pizzati_2020_WACV}
Fabio Pizzati, Raoul~de Charette, Michela Zaccaria, and Pietro Cerri.
\newblock Domain bridge for unpaired image-to-image translation and
  unsupervised domain adaptation.
\newblock In {\em WACV}, 2020.

\bibitem{roy2019DWT}
Subhankar Roy, Aliaksandr Siarohin, Enver Sangineto, Samuel~Rota Bulo, Nicu
  Sebe, and Elisa Ricci.
\newblock Unsupervised domain adaptation using feature-whitening and consensus
  loss.
\newblock In {\em Proceedings of the IEEE/CVF Conference on Computer Vision and
  Pattern Recognition}, 2019.

\bibitem{saito2020DANCE}
Kuniaki Saito, Donghyun Kim, Stan Sclaroff, and Kate Saenko.
\newblock Universal domain adaptation through self supervision.
\newblock {\em Advances in Neural Information Processing Systems}, 2020.

\bibitem{saito2018MCD}
Kuniaki Saito, Kohei Watanabe, Yoshitaka Ushiku, and Tatsuya Harada.
\newblock Maximum classifier discrepancy for unsupervised domain adaptation.
\newblock In {\em Proceedings of the IEEE/CVF Conference on Computer Vision and
  Pattern Recognition}, 2018.

\bibitem{sankaranarayanan2018GTA}
Swami Sankaranarayanan, Yogesh Balaji, Carlos~D Castillo, and Rama Chellappa.
\newblock Generate to adapt: Aligning domains using generative adversarial
  networks.
\newblock In {\em Proceedings of the IEEE/CVF Conference on Computer Vision and
  Pattern Recognition}, 2018.

\bibitem{shu2018dirt}
Rui Shu, Hung Bui, Hirokazu Narui, and Stefano Ermon.
\newblock A dirt-t approach to unsupervised domain adaptation.
\newblock In {\em International Conference on Learning Representations}, 2018.

\bibitem{sun2016DCORAL}
Baochen Sun and Kate Saenko.
\newblock Deep coral: Correlation alignment for deep domain adaptation.
\newblock In {\em Proceedings of the European Conference on Computer Vision},
  2016.

\bibitem{tang2020SRDC}
Hui Tang, Ke Chen, and Kui Jia.
\newblock Unsupervised domain adaptation via structurally regularized deep
  clustering.
\newblock In {\em Proceedings of the IEEE/CVF Conference on Computer Vision and
  Pattern Recognition}, pages 8725--8735, 2020.

\bibitem{tseng2019FWT}
Hung-Yu Tseng, Hsin-Ying Lee, Jia-Bin Huang, and Ming-Hsuan Yang.
\newblock Cross-domain few-shot classification via learned feature-wise
  transformation.
\newblock In {\em International Conference on Learning Representations}, 2019.

\bibitem{tzeng2017ADDA}
Eric Tzeng, Judy Hoffman, Kate Saenko, and Trevor Darrell.
\newblock Adversarial discriminative domain adaptation.
\newblock {\em CoRR}, abs/1702.05464, 2017.

\bibitem{tzeng2014deep}
Eric Tzeng, Judy Hoffman, Ning Zhang, Kate Saenko, and Trevor Darrell.
\newblock Deep domain confusion: Maximizing for domain invariance.
\newblock {\em arXiv preprint arXiv:1412.3474}, 2014.

\bibitem{tsne}
Laurens van~der Maaten and Geoffrey Hinton.
\newblock Visualizing data using t-sne.
\newblock {\em JMLR}, pages 2579--2605, 2008.

\bibitem{venkat2020SImpAl}
Naveen Venkat, Jogendra~Nath Kundu, Durgesh Singh, Ambareesh Revanur, and
  Venkatesh~Babu R.
\newblock Your classifier can secretly suffice multi-source domain adaptation.
\newblock In {\em Advances in Neural Information Processing Systems}, pages
  4647--4659, 2020.

\bibitem{venkateswara2017deep}
Hemanth Venkateswara, Jose Eusebio, Shayok Chakraborty, and Sethuraman
  Panchanathan.
\newblock Deep hashing network for unsupervised domain adaptation.
\newblock In {\em Proceedings of the IEEE/CVF Conference on Computer Vision and
  Pattern Recognition}, 2017.

\bibitem{wang2020SPL}
Qian Wang and Toby Breckon.
\newblock Unsupervised domain adaptation via structured prediction based
  selective pseudo-labeling.
\newblock In {\em Proceedings of the AAAI Conference on Artificial
  Intelligence}, pages 6243--6250, 2020.

\bibitem{wang2019TN}
Ximei Wang, Ying Jin, Mingsheng Long, Jianmin Wang, and Michael~I Jordan.
\newblock Transferable normalization: Towards improving transferability of deep
  neural networks.
\newblock In {\em Advances in Neural Information Processing Systems}, 2019.

\bibitem{wu2018GN}
Yuxin Wu and Kaiming He.
\newblock Group normalization.
\newblock In {\em Proceedings of the European Conference on Computer Vision},
  2018.

\bibitem{wu2020dual}
Yuan Wu, Diana Inkpen, and Ahmed El-Roby.
\newblock Dual mixup regularized learning for adversarial domain adaptation.
\newblock In {\em Proceedings of the European Conference on Computer Vision},
  2020.

\bibitem{xiao2021DWL}
Ni Xiao and Lei Zhang.
\newblock Dynamic weighted learning for unsupervised domain adaptation.
\newblock In {\em Proceedings of the IEEE/CVF Conference on Computer Vision and
  Pattern Recognition}, pages 15242--15251, 2021.

\bibitem{xu2020cross}
Minghao Xu, Hang Wang, Bingbing Ni, Qi Tian, and Wenjun Zhang.
\newblock Cross-domain detection via graph-induced prototype alignment.
\newblock In {\em Proceedings of the IEEE/CVF Conference on Computer Vision and
  Pattern Recognition}, 2020.

\bibitem{xu2019AFN}
Ruijia Xu, Guanbin Li, Jihan Yang, and Liang Lin.
\newblock Larger norm more transferable: An adaptive feature norm approach for
  unsupervised domain adaptation.
\newblock In {\em Proceedings of the IEEE International Conference on Computer
  Vision}, 2019.

\bibitem{yosinski2014transferable}
Jason Yosinski, Jeff Clune, Yoshua Bengio, and Hod Lipson.
\newblock How transferable are features in deep neural networks?
\newblock In {\em Advances in Neural Information Processing Systems}, 2014.

\bibitem{zhang2018IWAN}
Jing Zhang, Zewei Ding, Wanqing Li, and Philip Ogunbona.
\newblock Importance weighted adversarial nets for partial domain adaptation.
\newblock In {\em Proceedings of the IEEE/CVF Conference on Computer Vision and
  Pattern Recognition}, 2018.

\bibitem{zhang2018iCAN}
Weichen Zhang, Wanli Ouyang, Wen Li, and Dong Xu.
\newblock Collaborative and adversarial network for unsupervised domain
  adaptation.
\newblock In {\em Proceedings of the IEEE/CVF Conference on Computer Vision and
  Pattern Recognition}, 2018.

\bibitem{zhang2020APL}
Yabin Zhang, Bin Deng, Kui Jia, and Lei Zhang.
\newblock Label propagation with augmented anchors: A simple semi-supervised
  learning baseline for unsupervised domain adaptation.
\newblock In {\em Proceedings of the European Conference on Computer Vision},
  pages 781--797, 2020.

\bibitem{zhang2019MDD}
Yuchen Zhang, Tianle Liu, Mingsheng Long, and Michael Jordan.
\newblock Bridging theory and algorithm for domain adaptation.
\newblock In {\em International Conference on Machine Learning}, 2019.

\bibitem{zhou2020affinity}
Wei Zhou, Yukang Wang, Jiajia Chu, Jiehua Yang, Xiang Bai, and Yongchao Xu.
\newblock Affinity space adaptation for semantic segmentation across domains.
\newblock In {\em IEEE Transactions on Image Processing}. IEEE, 2020.

\bibitem{zou2019CRST}
Yang Zou, Zhiding Yu, Xiaofeng Liu, BVK Kumar, and Jinsong Wang.
\newblock Confidence regularized self-training.
\newblock In {\em Proceedings of the IEEE International Conference on Computer
  Vision}, 2019.

\bibitem{zuo2021attention}
Yukun Zuo, Hantao Yao, and Changsheng Xu.
\newblock Attention-based multi-source domain adaptation.
\newblock {\em IEEE Transactions on Image Processing}, 30:3793--3803, 2021.

\end{thebibliography}

\newpage
\begin{IEEEbiographynophoto}{Zhiyong Huang}
received his B.Eng. degree in Control and Computer Engineering from North China Electric Power University in July 2021. He received his B.Eng. degree from North China Electric Power University in 2018. His research interest include single image super-resolution and domain adaptation.
\end{IEEEbiographynophoto}

\begin{IEEEbiographynophoto}{Kekai Sheng}
received his PhD. degree from National Laboratory of Pattern Recognition (NLPR), Institute of Automation, Chinese Academy of Sciences in 2019. He received his B.Eng. degree in Telecommunication Engineering from University of Science and Technology Beijing in 2014. He is
currently a researcher engineer at Youtu Lab, Tencent Inc. His research interests include image quality evaluation, domain adaptation, and AutoML.
\end{IEEEbiographynophoto}

\begin{IEEEbiographynophoto}{Ke Li}
received the B.Eng. degree in Computer Science from Xiamen University, Fujian, China, in July 2018. He is currently a research engineer at Youtu Lab, Tencent Inc. His research interests involve self-supervised learning, deep learning, and machine learning.
\end{IEEEbiographynophoto}

\begin{IEEEbiographynophoto}{Jian Liang}
received the B.E. degree in Electronic Information and Technology from Xi'an Jiaotong University and Ph.D. degree in Pattern Recognition and Intelligent Systems from from National Laboratory of Pattern Recognition (NLPR), Institute of Automation, Chinese Academy of Sciences in July 2013, and January 2019, respectively. 
He was a research fellow at National University of Singapore from June 2019 to April 2021. 
Now he joins NLPR and works as an associated professor. 
His research interests focus on transfer learning, pattern recognition, and computer vision.
\end{IEEEbiographynophoto}

\begin{IEEEbiographynophoto}{Taiping Yao}
received the B.Eng. degree in Electrical Engineering from Shanghai Jiao Tong University, Shanghai, China, in July 2019. He is currently a researcher engineer at Youtu Lab, Tencent Inc. His research interests involve computer vision and deep learning.
\end{IEEEbiographynophoto}

\begin{IEEEbiographynophoto}{Weiming Dong}
is a Professor in the Sino-European Lab in Computer Science, Automation and Applied Mathematics (LIAMA) and National Laboratory of Pattern Recognition (NLPR) at Institute of Automation, Chinese Academy of Sciences. He received his BSc and MSc degrees in Computer Science in 2001 and 2004, both from Tsinghua University, China. He received his PhD in Computer Science from the University of Lorraine, France, in 2007. His research interests include visual media synthesis and image recognition. Weiming Dong is a member of the ACM and IEEE.
\end{IEEEbiographynophoto}


\begin{IEEEbiographynophoto}{Dengwen Zhou}
is a Professor in the School of Control and Computer Engineering, North China Electric Power University, Beijing, China. He has long been engaged in research on image processing, including image de-noising, image de-mosaicking, image interpolation and image super-resolution etc. Current research focuses on the applications based on neural networks and deep learning in image processing and computer vision.
\end{IEEEbiographynophoto}

\begin{IEEEbiographynophoto}{Xing Sun}
is currently a team lead and senior researcher in Youtu Lab, Tencent Inc. Before that, he received his Ph.D. degree under the supervision of Prof. Edmund Y. Lam in Imaging Systems Laboratory, and Dr. Nelson Yung in Laboratory for Intelligent Transportation Systems Research in the Department of Electrical and Electronic Engineering at The University of Hong Kong in 2016. He received his B.S. degree at Nanjing University of Science and Technology in Jun. 2012. 
\end{IEEEbiographynophoto}

\end{document}